\documentclass[lettersize,journal]{IEEEtran}

\usepackage{amsmath,amssymb,amsfonts}
\usepackage{algorithmic}
\usepackage{algorithm}
\usepackage{array}
\usepackage[caption=false,font=normalsize,labelfont=sf,textfont=sf]{subfig}
\usepackage{textcomp}
\usepackage{url}
\usepackage{verbatim}
\usepackage{graphicx}
\usepackage{cite}
\usepackage{bm}
\usepackage{multirow}
\usepackage{booktabs}
\usepackage{dblfloatfix}
\usepackage{microtype}
\usepackage{xcolor}
\graphicspath{{figures/}{./}}

\begin{document}

\title{Attribution-Guided Multimodal Deepfake Detection\\via Cross-Modal Forensic Fingerprints}

\author{Wasim Ahmad, Wei Zhang and Xuerui Mao~\thanks{*Corresponding author: Xuerui Mao}\\
\IEEEauthorblockA{Wasim Ahmad, Wei Zhang, and Xuerui Mao are with the School of Interdisciplinary Science, Beijing Institute of Technology, Beijing 100081, China (Emails: 7520250351@bit.edu.cn, w.w.zhanger@gmail.com, xmao@bit.edu.cn)}\\
% \IEEEauthorblockA{Yan-Tsung Peng is with the Department of Computer Science, National Chengchi University, 116, Taipei, Taiwan (Email: ytpeng@cs.nccu.edu.tw)}
}

% The paper headers
% \markboth{IEEE Transactions on Information Forensics and Security}%
% {Ahmad \MakeLowercase{\textit{et al.}}: Attribution-Guided Multimodal Deepfake Detection}

% \IEEEpubid{0000--0000/00\$00.00~\copyright~2025 IEEE}

\maketitle

\setcounter{dbltopnumber}{4}
\setcounter{topnumber}{4}
\setcounter{totalnumber}{6}
\renewcommand{\floatpagefraction}{0.9}
\renewcommand{\dblfloatpagefraction}{0.9}
\renewcommand{\dbltopfraction}{0.95}
\renewcommand{\topfraction}{0.95}
\renewcommand{\textfraction}{0.05}

\begin{figure*}[t]
    \centering
    \includegraphics[width=\textwidth]{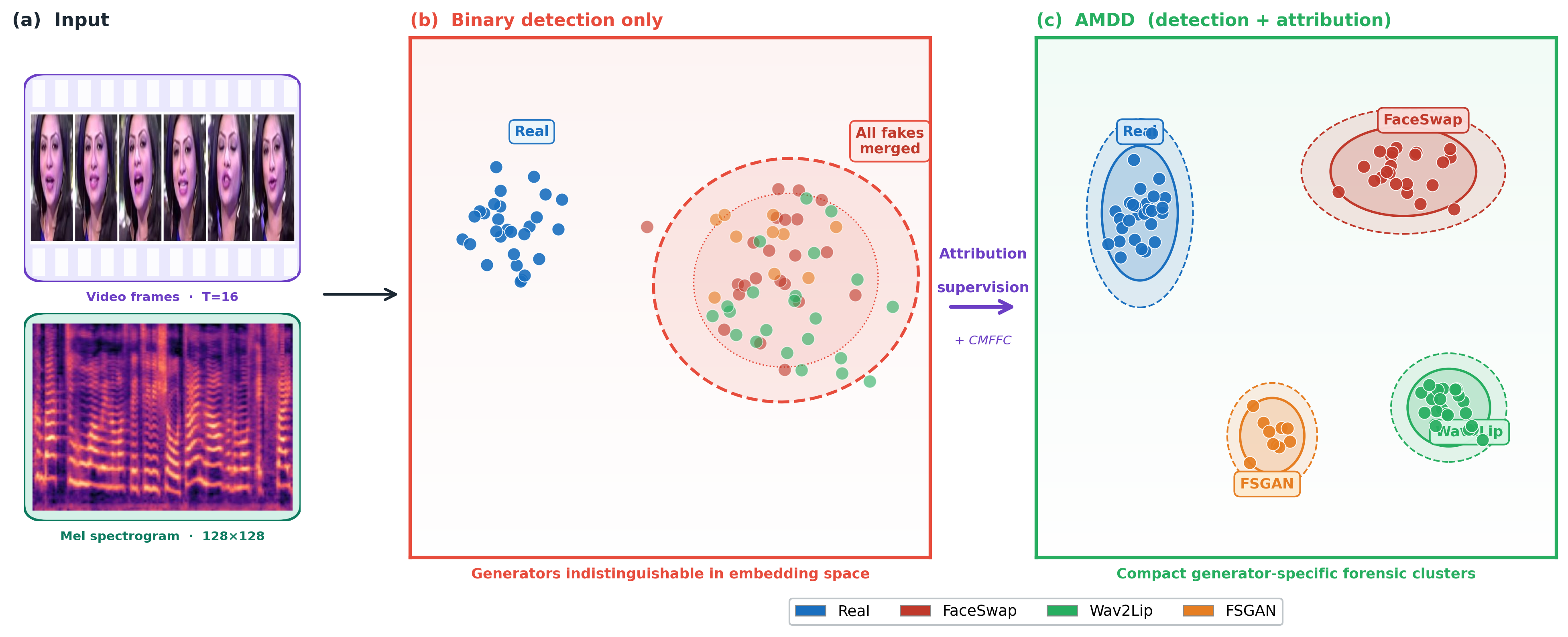}
    \caption{Motivation for attribution-guided detection. (a) Real video frames and mel spectrogram serve as multimodal inputs. (b) A detector trained with a binary objective only collapses all fake samples into an unstructured embedding region regardless of source generator, learning shallow shortcuts rather than forensic fingerprints. (c) AMDD, jointly trained with attribution supervision and the CMFFC loss, produces compact generator-specific clusters while maintaining clear real/fake separation, demonstrating that attribution-guided learning produces qualitatively richer forensic representations.}
    \label{fig:teaser}
\end{figure*}

% ==================================================
\begin{abstract}
Audio-visual deepfakes have reached a level of realism that makes perceptual detection unreliable, creating serious risks to media integrity, biometric security, and political discourse. Although multimodal detection approaches have shown promise by combining audio and visual cues, they remain largely formulated as binary classification tasks optimized purely for authenticity prediction. This narrow objective encourages models to latch onto dataset-specific artifacts rather than the underlying generative traces that are genuinely distinctive across manipulation methods. We argue that a detector which cannot identify \textit{how} a video was forged is inevitably learning the wrong signal --- unlike attribution-guided learning, which imposes a strictly stronger geometric constraint on the shared embedding space than binary detection alone, forcing the model to encode generator-specific forensic content rather than dataset-specific shortcuts. To address this, we propose the Attribution-Guided Multimodal Deepfake Detection (AMDD) framework, which jointly learns to detect and attribute manipulation within a shared embedding space. The core idea is to treat generator attribution (identifying which deepfake model produced a given sample) as structured regularization that constrains the representation geometry toward forensically meaningful features. We further introduce a Cross-Modal Forensic Fingerprint Consistency (CMFFC) loss that explicitly enforces alignment between generator-induced artifacts in the visual and audio streams, exploiting the observation that coherent manipulation leaves correlated traces across modalities, a signal fundamentally grounded in the physical coupling between speech production and facial articulation that authentic recordings exhibit and that synthetic pipelines routinely disrupt. Architecturally, we pair a ResNet50 with temporal attention for visual encoding against a pretrained ResNet18 adapted for mel spectrograms, deliberately closing the encoder capacity gap that causes prior multimodal detectors to underutilize audio. On FakeAVCeleb, AMDD achieves 99.7\% balanced accuracy and 99.8\% AUC with 95.9\% attribution accuracy. Cross-dataset evaluation on DeepfakeTIMIT, DFDM, and LAV-DF confirms that real video detection generalizes robustly, while fake detection on unseen generators remains an open challenge that we analyze in depth.
\end{abstract}


% ==================================================
\section{Introduction}
\label{sec:intro}

\IEEEPARstart{S}{ynthetic} media generation has matured to a point where fabricated audio-visual content (faces convincingly swapped, lips synchronized to arbitrary speech, voices cloned from seconds of audio) is within reach of anyone with consumer hardware. Methods such as FaceSwap~\cite{faceswap}, FSGAN~\cite{fsgan}, Wav2Lip~\cite{wav2lip}, and neural voice synthesis~\cite{sv2tts} have closed the gap between professional production and amateur forgery, making perceptually convincing deepfakes a routine rather than exceptional threat. The downstream consequences are well documented: political disinformation, identity fraud, biometric spoofing, and the broader corrosion of trust in video as evidence~\cite{fakeavceleb}.

Detection has responded in kind, with multimodal approaches gaining traction by combining visual and audio streams to improve robustness over single-modality baselines~\cite{jointAV,mds,emotions}. The intuition is sound: a convincing deepfake must maintain audio-visual coherence, and failure to do so should be detectable. Yet most existing methods reduce this rich forensic problem to a binary classification task, training a model to output ``real'' or ``fake'' and nothing more. The consequence is well known in the domain adaptation literature: when the decision boundary is not grounded in causal features, models tend to exploit whatever statistical regularities are most convenient in the training distribution: compression artifacts, background noise patterns, identity-linked cues, none of which generalize across datasets or generators~\cite{cmaldd,npvforensics}.

The forensics community has long recognized that generative models are not interchangeable: each leaves a characteristic signature in the content it produces, detectable in frequency statistics~\cite{ganfingerprint}, phase patterns, and temporal dynamics. These signatures persist across samples from the same generator and differ systematically between generators. If a detector cannot tell a Wav2Lip forgery from a FaceSwap forgery, it is unlikely to have learned these signatures at all. It has instead learned something more shallow. This observation motivates the central question of this paper: can making a detector answer the harder question of \textit{which} generator produced a sample actually make it better at the easier question of \textit{whether} a sample was generated at all? Fig.~\ref{fig:teaser} illustrates this motivation directly: a binary detector produces unstructured embeddings that conflate all generators, while AMDD produces compact generator-specific clusters, confirming that attribution supervision drives the model toward forensically meaningful representations.

Our prior work on visual attribution provides empirical grounding for a positive answer. FAME~\cite{fame2025} demonstrated that a lightweight spatio-temporal network trained purely for attribution achieves strong generator identification from visual features alone, implying that those features carry genuine forensic information. CapST~\cite{capst2025} subsequently showed that capsule-based temporal attention improves attribution further by preserving equivariant spatial structure. Ahmad et al.~\cite{resvit} explored residual-transformer hybrids for deepfake detection, and Hashmi et al.~\cite{hashmi2022} found that ensemble strategies improve multimodal forgery detection. None of these works, however, address the question of whether attribution can \textit{regularize} detection, or whether cross-modal attribution is feasible or useful at all.

AMDD answers both questions. We jointly train detection and attribution within a single embedding space, using attribution supervision as a structured regularizer that penalizes representations which cannot discriminate generators. We further introduce a Cross-Modal Forensic Fingerprint Consistency (CMFFC) loss that pulls the visual and audio representations of same-generator samples toward alignment, formalizing the intuition that a generator's trace should be coherent across modalities. The architecture pairs a ResNet50 with temporal attention for video encoding against a pretrained ResNet18 adapted for mel spectrograms, a design choice driven by the observation that prior multimodal detectors use audio encoders an order of magnitude smaller than their visual counterparts, effectively discarding the audio signal before fusion.

The contributions of this work are summarized as follows:

\begin{itemize}
    \item We propose AMDD, a multimodal deepfake detection framework that jointly trains detection and generator attribution within a single embedding space. We show that attribution supervision functions as structured regularization, shaping the representation geometry toward forensically meaningful features even when binary detection accuracy is already near-ceiling.

    \item We introduce the Cross-Modal Forensic Fingerprint Consistency (CMFFC) loss, which enforces alignment between the visual and audio representations of same-generator samples. This formalizes the observation that a coherent manipulation leaves correlated traces across modalities, providing a training signal that no prior multimodal detector has exploited.

    \item We close the encoder capacity gap present in existing multimodal detectors by adapting a full pretrained ResNet18 for audio encoding rather than the shallow CNNs typically used, pairing it against a ResNet50 with temporal attention for video encoding.

    \item We conduct cross-dataset evaluation on DeepfakeTIMIT, DFDM, and LAV-DF and provide a detailed analysis of where generalization succeeds (real video detection transfers robustly) and where it fails (fake detection on unseen generators does not), along with a mechanistic explanation of why.
\end{itemize}

% ==================================================
\section{Related Work}
\label{sec:related}

\subsection{Visual Deepfake Detection}

Detection of facial forgeries began with physiological signals that early generators struggled to replicate. Li et al.~\cite{eyeblinking} observed that synthesized faces rarely blinked at natural frequencies, providing a reliable early cue, while Yang et al.~\cite{headpose} showed that head pose estimates derived from 3D landmarks expose geometric inconsistencies introduced by face alignment in swap pipelines. These handcrafted signals were quickly obsoleted as generation quality improved, motivating a shift toward learned representations. XceptionNet~\cite{xception} emerged as a standard backbone for frame-level classification, benefiting from its strong inductive bias toward high-frequency texture features that forgeries tend to alter. Temporal reasoning was subsequently incorporated through recurrent architectures~\cite{lstm} and 3D convolutions~\cite{3dcnn}, allowing models to exploit inter-frame inconsistencies that frame-level detectors miss. FaceForensics++~\cite{ffpp} provided the community with a large-scale benchmark spanning four manipulation types, and most subsequent detection methods report results on this dataset, making it the de facto standard for comparison.

More recent work has moved away from global classification toward spatially and spectrally informed representations. LipForensics~\cite{lipforensics} fine-tunes a lip-reading network for detection, leveraging the observation that face-swap methods distort the fine-grained mouth region dynamics that speech requires. RECCE~\cite{recce} reframes detection as an anomaly problem, using reconstruction error from an identity-preserving autoencoder to surface traces that a discriminative classifier might overlook. Frank et al.~\cite{frequency} demonstrated that GAN-generated content carries systematic high-frequency artifacts invisible in the spatial domain, and spectral analysis remains a productive complementary signal. Zhao et al.~\cite{sst} proposed a self-supervised transformer that learns forgery-discriminative features through masked reconstruction, showing that pretext tasks can surface forensic structure without explicit labels. Ahmad et al.~\cite{resvit} integrated residual convolutional learning with vision transformers for video-level detection. Despite these advances, visual-only detectors share a structural limitation: they are blind to audio-channel manipulation and, in practice, tend to overfit to the specific generator families present in training data.

\subsection{Audio Deepfake Detection}

Speech synthesis and voice conversion have progressed in parallel with visual forgery, producing synthetic speech that is increasingly indistinguishable from genuine recordings. The ASVspoof challenge series~\cite{asvspoof} has been the primary benchmark driving progress in this area, providing large-scale controlled evaluations across a range of synthesis and conversion conditions. Strong performing systems include RawNet~\cite{rawnet}, which operates on raw waveforms without hand-engineered features, and AASIST~\cite{aasist}, which combines spectral and temporal graph attention to model the multi-scale artifact structure introduced by modern vocoders. Both architectures have demonstrated that end-to-end learning from audio alone can achieve high accuracy within their training distribution.

The fundamental limitation of audio-only approaches, however, is scope: they are designed and evaluated in settings where video is unavailable or irrelevant. This makes them poorly suited to the increasingly common scenario where a video's visual content has been manipulated while the audio track remains authentic, or vice versa. Detecting such asymmetric forgeries requires reasoning about the relationship between modalities, not just the properties of each in isolation. This cross-modal reasoning is precisely what multimodal detection methods attempt to provide.

\subsection{Multimodal Deepfake Detection}

Combining audio and visual evidence has become an active research direction motivated by the complementary nature of the two modalities: a face-swap that leaves audio intact will exhibit visual artifacts absent from the audio, while audio synthesis paired with real video will show the reverse pattern. Chugh et al.~\cite{mds} were among the first to formalize this intuition, introducing a contrastive formulation that penalizes audio-visual dissonance and can localize temporal segments where the two streams become inconsistent. Mittal et al.~\cite{emotions} pursued a semantically richer signal, using the emotional congruence between facial expression and vocal affect as a forensic cue: genuine recordings tend to maintain affective consistency across modalities, while fabrications often do not. The Joint Audio-Visual framework~\cite{jointAV} proposed synchronized dual-stream encoding with cross-modal attention, establishing a practical baseline for feature-level fusion in deepfake detection.

Subsequent work has pushed performance considerably higher through more powerful backbone architectures and more sophisticated fusion strategies. CMALDD-PTAF~\cite{cmaldd} exploits pre-trained audio-visual representations with cross-attention fusion to reach 99.6\% AUC on FakeAVCeleb; NPVForensics~\cite{npvforensics} targets the alignment between phoneme sequences and their corresponding viseme regions, a signal grounded in the physical constraints of speech production that fabrications frequently violate; and ERF-BA-TFD+~\cite{erfba} combines enhanced receptive fields with uncertainty-based evidence reasoning to improve detection under challenging conditions. AVForensics~\cite{avforensics} introduced a masking-based training strategy that randomly suppresses audio or visual inputs during training, forcing the model to develop robust representations that do not depend entirely on either modality. Hashmi et al.~\cite{hashmi2022} demonstrated that ensemble strategies across complementary multimodal detectors consistently outperform any individual component.

Despite this progress, a common thread runs through all of these approaches: they optimize a binary detection objective, training the model to output ``real'' or ``fake'' without any supervision on which generator was responsible. The representations that emerge from this training may be effective within a given dataset, but they are not necessarily anchored to the forensic traces that distinguish different generative processes. Our work is motivated by the hypothesis that imposing this additional structure, through joint attribution supervision, will produce representations that are both more interpretable and more robust.

\subsection{Deepfake Attribution}

The question of which model produced a synthetic sample has received comparatively little attention relative to the binary detection problem. The foundational insight comes from Wang et al.~\cite{ganfingerprint}, who demonstrated that CNN-generated images are surprisingly easy to attribute to their source architecture: generators leave distinctive frequency-domain fingerprints in their outputs, consistent across samples and discriminative across different networks. This result implies that the information needed to attribute a forgery to its generator is genuinely present in the signal, and that it persists even after compression and resizing. Verdoliva~\cite{videofingerprint} provided a broader survey of video forensics, situating attribution within the larger landscape of manipulation detection and noting that temporal aggregation of frame-level fingerprints can substantially improve reliability at the video level. Jia et al.~\cite{DMASTA2022} specifically tackled model attribution for face-swap deepfakes in video, demonstrating that generator-specific traces persist across frames and that temporal aggregation strategies significantly improve attribution accuracy compared to frame-level classification alone.

Similar fingerprinting behavior has been documented in the audio domain~\cite{audiofingerprint}, suggesting that the phenomenon is general across synthesis modalities rather than specific to image generation. Our own prior work has built on these foundations from a practical attribution perspective: FAME~\cite{fame2025} showed that a compact spatio-temporal network can reliably identify the source face-swap generator from visual features alone, and CapST~\cite{capst2025} improved on this using capsule networks with temporal attention, which better preserve the equivariant spatial structure needed to distinguish manipulation styles that differ primarily in their blending boundary characteristics.

The present work departs from this line of attribution-as-end-task research by treating attribution not as a standalone end task but as a training-time regularizer that shapes the detection embedding space. The distinction matters: a standalone attribution model optimizes for generator identification independently and produces no direct benefit for detection. By contrast, when attribution supervision is imposed jointly within a shared embedding space, it constrains what the representation must encode. Binary detection requires only that real and fake embeddings be linearly separable, a weak constraint that many shallow representations can satisfy. Attribution requires additionally that embeddings from different generators be mutually separable, a strictly stronger constraint that forces the model to encode generator-specific forensic content. The detection objective is subsumed by the attribution objective in this sense: any representation that separates generators will also separate real from fake, but not vice versa. Attribution therefore acts as a principled regularizer that guides the shared representation toward forensically structured geometry, rather than simply adding a second output head.

% ==================================================
\section{Proposed Framework}
\label{sec:method}

\subsection{Problem Formulation}

Let $\mathbf{v} = \{f_1, f_2, \ldots, f_T\}$ denote a sequence of $T$ video frames and $\mathbf{a}$ denote the corresponding mel spectrogram extracted from the audio track. Given an audio-visual input pair $(\mathbf{v}, \mathbf{a})$, our framework simultaneously predicts: (i) a binary detection label $y \in \{0, 1\}$ indicating real ($y=0$) or fake ($y=1$); and (ii) a generator attribution label $g \in \{0, 1, \ldots, G\}$ where $g=0$ denotes a real sample and $g \in \{1, \ldots, G\}$ denotes one of $G$ known fake generators.

The core hypothesis is that jointly optimizing for both objectives constrains the shared embedding space to capture genuine forensic fingerprints rather than dataset-specific shortcuts, thereby improving generalization.

\subsection{Framework Overview}

\begin{figure*}[!t]
    \centering
    \includegraphics[width=\textwidth, trim=0 160 0 0, clip]{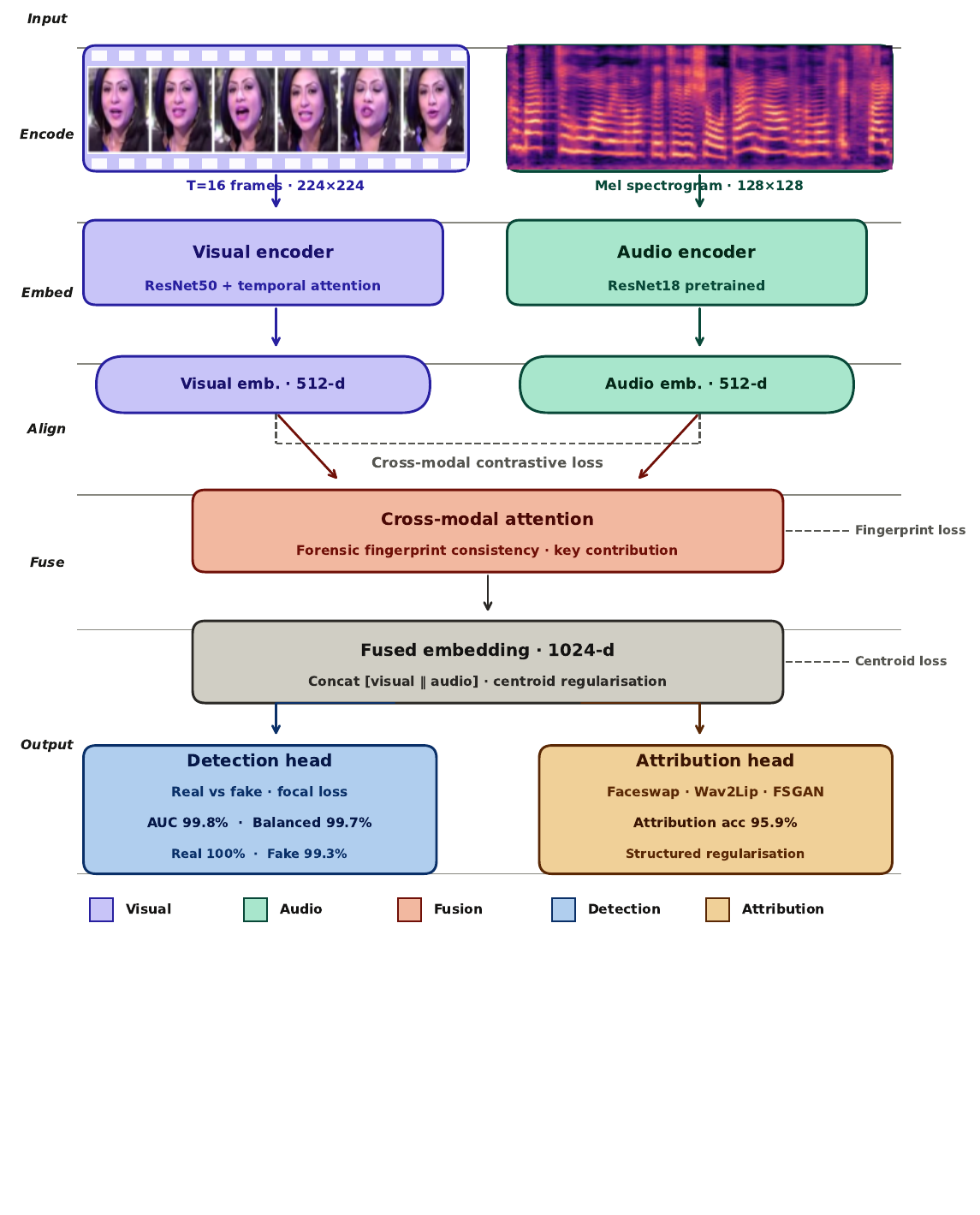}
    \caption{Overview of the proposed Attribution-Guided Multimodal Deepfake Detection (AMDD) framework. The visual stream encodes $T=16$ face-cropped frames via ResNet50 with temporal attention. The audio stream encodes a mel spectrogram via a pretrained ResNet18. Cross-modal attention aligns generator-induced forensic fingerprints across modalities. The fused 1024-d embedding is passed to a detection head (real vs.\ fake) and an attribution head (generator identity), jointly optimized with five loss functions.}
    \label{fig:framework}
\end{figure*}

The AMDD framework consists of five main components: a visual encoder $\mathcal{E}_v$, an audio encoder $\mathcal{E}_a$, a cross-modal attention module $\mathcal{A}$, a detection head $\mathcal{H}_d$, and an attribution head $\mathcal{H}_g$. The overall architecture is illustrated in Fig.~\ref{fig:framework}.

Given inputs $(\mathbf{v}, \mathbf{a})$, the framework produces:
\begin{equation}
    \mathbf{z}_v = \mathcal{E}_v(\mathbf{v}), \quad \mathbf{z}_a = \mathcal{E}_a(\mathbf{a})
\end{equation}
\begin{equation}
    \tilde{\mathbf{z}}_v, \tilde{\mathbf{z}}_a = \mathcal{A}(\mathbf{z}_v, \mathbf{z}_a)
\end{equation}
\begin{equation}
    \mathbf{z}_f = [\tilde{\mathbf{z}}_v \| \tilde{\mathbf{z}}_a] \in \mathbb{R}^{2D}
\end{equation}

where $[\cdot \| \cdot]$ denotes concatenation and $D=512$ is the embedding dimension. The fused embedding $\mathbf{z}_f$ is then passed to both heads:
\begin{equation}
    \hat{y} = \sigma(\mathcal{H}_d(\mathbf{z}_f)), \quad \hat{g} = \text{softmax}(\mathcal{H}_g(\mathbf{z}_f))
\end{equation}

\subsection{Visual Encoder}

The visual encoder processes a sequence of $T=16$ face-cropped frames. Each frame is passed through a pretrained ResNet50 backbone~\cite{resnet} with the classification head removed, producing a 2048-dimensional feature vector per frame. These features are projected to a $D$-dimensional space via a linear layer followed by layer normalization and ReLU activation. The resulting frame feature sequence $\{\mathbf{h}_1, \ldots, \mathbf{h}_T\} \in \mathbb{R}^{T \times D}$ is then processed by a temporal attention module:

\begin{equation}
    \text{Attn}(\mathbf{H}) = \text{LayerNorm}(\mathbf{H} + \text{MHA}(\mathbf{H}, \mathbf{H}, \mathbf{H}))
\end{equation}

where MHA denotes multi-head attention with 8 heads. The video embedding $\mathbf{z}_v \in \mathbb{R}^D$ is obtained by mean pooling over the temporal dimension. This design captures both local appearance artifacts in individual frames and temporal inconsistencies across the sequence.

\subsection{Audio Encoder}

A key architectural contribution of AMDD is the balanced audio encoder design. Prior multimodal detectors typically employ lightweight audio networks (e.g., 3-layer CNNs) paired with deep visual backbones, creating an encoder capacity imbalance that causes the model to implicitly ignore audio features. We address this by adapting a pretrained ResNet18 for mel spectrogram processing. The first convolutional layer is modified to accept single-channel input by averaging the pretrained RGB weights across channels, preserving ImageNet initialization. The classification head is replaced with a projection layer mapping to the same $D$-dimensional space as the visual encoder. This design provides 11.4M parameters for audio encoding versus 25.6M for visual encoding, a substantially more balanced configuration than prior work, as summarized in Table~\ref{tab:arch}.

Audio input is a $128 \times 128$ mel spectrogram computed from 4 seconds of audio at 16 kHz with 1024-point FFT and hop length of 512, converted to decibels and normalized to $[-1, 1]$. Mel spectrogram characteristics across manipulation types are analyzed in Section~\ref{sec:discussion}.

\subsection{Cross-Modal Attention}

The cross-modal attention module $\mathcal{A}$ enables each modality to attend to the other, aligning generator-induced artifacts across streams. Given visual embedding $\mathbf{z}_v$ and audio embedding $\mathbf{z}_a$:

\begin{equation}
    \tilde{\mathbf{z}}_v = \text{LayerNorm}\left(\mathbf{z}_v + \text{MHA}(\mathbf{z}_v, \mathbf{z}_a, \mathbf{z}_a)\right)
\end{equation}
\begin{equation}
    \tilde{\mathbf{z}}_a = \text{LayerNorm}\left(\mathbf{z}_a + \text{MHA}(\mathbf{z}_a, \mathbf{z}_v, \mathbf{z}_v)\right)
\end{equation}

This bidirectional cross-attention allows the model to identify correlated artifacts that a generator introduces simultaneously in both modalities. For instance, a face-swap generator that replaces facial regions will alter both lip motion patterns (visual) and the alignment between facial movements and speech (audio).

\subsection{Detection and Attribution Heads}

Both heads are implemented as two-layer MLPs with ReLU activation and dropout. The detection head produces a scalar logit for binary classification. The attribution head produces a $(G+1)$-dimensional logit vector where the first class corresponds to real samples. Table~\ref{tab:arch} provides a complete summary of all components and their parameter counts.

\begin{table}[t]
\centering
\caption{Architecture Summary of the AMDD Framework.}
\label{tab:arch}
\begin{tabular}{llcc}
\toprule
Component & Backbone & Params & Output Dim \\
\midrule
Visual encoder     & ResNet50~\cite{resnet} & 24.6M & 512-d \\
Temporal attention & MHA (8 heads)          & 1.0M  & 512-d \\
Audio encoder      & ResNet18~\cite{resnet} & 11.2M & 512-d \\
Cross-modal attn   & MHA (8 heads)          & 2.1M  & 512-d each \\
Detection head     & 2-layer MLP            & 0.3M  & 1 (logit) \\
Attribution head   & 2-layer MLP            & 0.3M  & 4 (classes) \\
Projection heads   & 2-layer MLP            & 0.5M  & 512-d each \\
\midrule
\textbf{Total}     & --                     & \textbf{40.7M} & -- \\
\bottomrule
\end{tabular}
\end{table}

\subsection{Loss Functions}

The total training loss combines five objectives:

\begin{equation}
    \mathcal{L} = \mathcal{L}_{\text{det}} + \lambda_a \mathcal{L}_{\text{attr}} + \lambda_c \mathcal{L}_{\text{cont}} + \lambda_f \mathcal{L}_{\text{fp}} + \lambda_r \mathcal{L}_{\text{cen}}
\end{equation}

\textbf{Detection loss} $\mathcal{L}_{\text{det}}$ uses focal loss~\cite{focalloss} to handle class imbalance:
\begin{equation}
    \mathcal{L}_{\text{det}} = -\alpha_t (1 - p_t)^\gamma \log(p_t)
\end{equation}
with $\alpha=0.75$ and $\gamma=2.0$.

\textbf{Attribution loss} $\mathcal{L}_{\text{attr}}$ applies cross-entropy over the $(G+1)$ generator classes:
\begin{equation}
    \mathcal{L}_{\text{attr}} = -\sum_{k=0}^{G} \mathbf{1}[g=k] \log \hat{g}_k
\end{equation}

\textbf{Cross-modal contrastive loss} $\mathcal{L}_{\text{cont}}$ aligns visual and audio projections from the same video:
\begin{equation}
    \mathcal{L}_{\text{cont}} = \frac{1}{2}\left(\mathcal{L}_{\text{NCE}}(\mathbf{p}_v, \mathbf{p}_a) + \mathcal{L}_{\text{NCE}}(\mathbf{p}_a, \mathbf{p}_v)\right)
\end{equation}
where $\mathbf{p}_v = \text{normalize}(\text{Proj}_v(\tilde{\mathbf{z}}_v))$ and $\mathbf{p}_a = \text{normalize}(\text{Proj}_a(\tilde{\mathbf{z}}_a))$ are $\ell_2$-normalized projections, and $\mathcal{L}_{\text{NCE}}$ is the InfoNCE loss with temperature $\tau=0.07$.

\textbf{Cross-Modal Forensic Fingerprint Consistency} $\mathcal{L}_{\text{fp}}$ is our key novel loss. For fake samples from the same generator $g$, it enforces alignment between visual and audio projections:
\begin{equation}
    \mathcal{L}_{\text{fp}} = \frac{1}{|G_{\text{fake}}|} \sum_{g \in G_{\text{fake}}} \mathcal{L}_{\text{NCE}}^{(g)}(\mathbf{p}_v^{(g)}, \mathbf{p}_a^{(g)})
\end{equation}
where $\mathbf{p}_v^{(g)}$ and $\mathbf{p}_a^{(g)}$ are projections of samples from generator $g$. This loss enforces the principle that a specific generator leaves coherent, cross-modally consistent fingerprints.

\textbf{Centroid regularization} $\mathcal{L}_{\text{cen}}$ maintains generator-specific prototype embeddings updated via exponential moving average with momentum $m=0.9$, pulling embeddings toward their generator centroid to stabilize the representation geometry.

Loss weights are set as listed in Table~\ref{tab:hyperparams}.

\subsection{Training Procedure}

Algorithm~\ref{alg:training} summarizes the complete AMDD training procedure. The model is initialized with pretrained ImageNet weights for both encoders, and all five loss components are computed jointly in each forward pass. Generator centroids are updated via EMA after each batch to maintain stable prototype representations throughout training.

\begin{algorithm}[t]
\caption{AMDD Training Procedure}
\label{alg:training}
\begin{algorithmic}[1]
\REQUIRE Training set $\mathcal{D} = \{(\mathbf{v}_i, \mathbf{a}_i, y_i, g_i)\}_{i=1}^{N}$, epochs $E$, loss weights $\lambda_a, \lambda_c, \lambda_f, \lambda_r$
\ENSURE Trained model $\theta$
\STATE Initialize $\mathcal{E}_v$ with ResNet50 (ImageNet pretrained)
\STATE Initialize $\mathcal{E}_a$ with ResNet18 (ImageNet, channel-averaged)
\STATE Initialize $\mathcal{A}$, $\mathcal{H}_d$, $\mathcal{H}_g$, centroids $\mathbf{C} \leftarrow \mathbf{0}$
\FOR{epoch $e = 1$ to $E$}
    \FOR{each mini-batch $\mathcal{B} \sim \mathcal{D}$}
        \STATE $\mathbf{z}_v \leftarrow \mathcal{E}_v(\mathbf{v})$, \quad $\mathbf{z}_a \leftarrow \mathcal{E}_a(\mathbf{a})$
        \STATE $\tilde{\mathbf{z}}_v, \tilde{\mathbf{z}}_a \leftarrow \mathcal{A}(\mathbf{z}_v, \mathbf{z}_a)$
        \STATE $\mathbf{z}_f \leftarrow [\tilde{\mathbf{z}}_v \| \tilde{\mathbf{z}}_a]$
        \STATE $\hat{y} \leftarrow \sigma(\mathcal{H}_d(\mathbf{z}_f))$, \quad $\hat{g} \leftarrow \text{softmax}(\mathcal{H}_g(\mathbf{z}_f))$
        \STATE $\mathbf{p}_v \leftarrow \text{normalize}(\text{Proj}_v(\tilde{\mathbf{z}}_v))$
        \STATE $\mathbf{p}_a \leftarrow \text{normalize}(\text{Proj}_a(\tilde{\mathbf{z}}_a))$
        \STATE Compute $\mathcal{L}_{\text{det}}$ via focal loss on $(\hat{y}, y)$
        \STATE Compute $\mathcal{L}_{\text{attr}}$ via cross-entropy on $(\hat{g}, g)$
        \STATE Compute $\mathcal{L}_{\text{cont}}$ via InfoNCE on $(\mathbf{p}_v, \mathbf{p}_a)$
        \STATE Compute $\mathcal{L}_{\text{fp}}$ via within-generator cross-modal NCE
        \STATE Compute $\mathcal{L}_{\text{cen}}$ via distance to centroids $\mathbf{C}$
        \STATE $\mathcal{L} \leftarrow \mathcal{L}_{\text{det}} + \lambda_a\mathcal{L}_{\text{attr}} + \lambda_c\mathcal{L}_{\text{cont}} + \lambda_f\mathcal{L}_{\text{fp}} + \lambda_r\mathcal{L}_{\text{cen}}$
        \STATE Update $\theta \leftarrow \theta - \eta \nabla_\theta \mathcal{L}$ with gradient clipping
        \STATE \textbf{EMA:} $\mathbf{C}[g] \leftarrow m \cdot \mathbf{C}[g] + (1-m) \cdot \bar{\mathbf{z}}_f^{(g)}$
    \ENDFOR
    \STATE Decay learning rate via cosine annealing
\ENDFOR
\RETURN $\theta$
\end{algorithmic}
\end{algorithm}

% ==================================================
\section{Experiments}
\label{sec:exp}

\subsection{Datasets}

Table~\ref{tab:datasets} summarizes all datasets used in this work.

\begin{table}[t]
\centering
\caption{Summary of Datasets Used in This Work. A+V = audio and video. V = video only.}
\label{tab:datasets}
\begin{tabular}{llcccc}
\toprule
Dataset & Split & Real & Fake & Modality & Audio \\
\midrule
\multirow{2}{*}{FakeAVCeleb~\cite{fakeavceleb}} & Train & 670  & 674  & A+V & \checkmark \\
 & Test  & 75   & 145  & A+V & \checkmark \\
\multirow{2}{*}{DF-TIMIT~\cite{dftimit}} & Train (real) & 320 & -- & A+V & \checkmark \\
 & Test  & 320  & 320  & A+V & \checkmark \\
DFDM~\cite{dfdm}     & Test  & 130  & 520  & V only & $\times$ \\
LAV-DF~\cite{lavdf}  & Test (sampled) & 500 & 498 & A+V & \checkmark \\
\bottomrule
\end{tabular}
\end{table}

\textbf{FakeAVCeleb}~\cite{fakeavceleb} is the primary training and evaluation dataset. It contains audio-visual deepfakes generated using FaceSwap, FSGAN, and Wav2Lip applied to VoxCeleb celebrities. We use the FakeVideo-RealAudio subset (Category C) as fake samples. To address the severe real/fake imbalance (approximately 1 real video per identity vs.\ 10--40 fakes), we supplement real training data with DF-TIMIT real identities, yielding 820 real and 964 fake samples after per-generator capping. Table~\ref{tab:gen_dist} details the split distribution.

Other widely used benchmarks include FaceForensics++~\cite{ffpp}, which provides frame-level forgeries across four manipulation methods, and Celeb-DF~\cite{celepdf}, which targets high-quality celebrity deepfakes that are harder to detect. The Deepfake Detection Challenge dataset~\cite{dolhansky2020} introduced large-scale evaluation under diverse real-world recording conditions. DeepfakeTIMIT~\cite{timit} is based on the original TIMIT corpus and provides controlled GAN-based face-swap forgeries at two quality levels.

\begin{table}[t]
\centering
\caption{Generator Distribution Across Splits After Per-Generator Capping.}
\label{tab:gen_dist}
\begin{tabular}{llccc}
\toprule
Generator & Type & Train & Val & Test \\
\midrule
Real (FakeAVCeleb)  & --     & 350  & 75  & 75 \\
Real (DF-TIMIT)     & --     & 320  & --  & -- \\
FaceSwap            & Visual & 265  & 57  & 57 \\
Wav2Lip             & Visual & 350  & 75  & 75 \\
FSGAN               & Visual & 59   & 13  & 13 \\
\midrule
Total               & --     & 1344 & 220 & 220 \\
\bottomrule
\end{tabular}
\end{table}

\textbf{DeepfakeTIMIT}~\cite{dftimit} provides GAN-based face-swap deepfakes in higher and lower quality variants with original speech audio stored separately per utterance. Used for cross-dataset evaluation only.

\textbf{DFDM}~\cite{dfdm} provides deepfakes generated by five methods at three quality levels (crf0, crf10, crf23). As DFDM contains no audio, it is used for visual-stream cross-dataset evaluation with silent audio as fallback.

\textbf{LAV-DF}~\cite{lavdf} is a large-scale multimodal forgery dataset with video-only, audio-only, and both-modality manipulated samples. We evaluate on a balanced 998-sample subset of the test split, stratified across manipulation types.

\subsection{Implementation Details}

All models are trained using AdamW~\cite{adamw} with learning rate $10^{-4}$, weight decay $10^{-4}$, and cosine annealing over 50 epochs. Batch size is 4 with gradient clipping at norm 1.0. Visual frames are resized to $224 \times 224$ with ImageNet normalization. Training augmentation includes random horizontal flip, color jitter, and random grayscale. A weighted random sampler balances generator classes during training. Face crops are extracted using face-alignment~\cite{facealignment} and cached as preprocessed numpy arrays. All experiments are conducted on a single NVIDIA RTX 5060 Ti (16GB VRAM). Table~\ref{tab:hyperparams} lists all hyperparameters.

\begin{table}[t]
\centering
\caption{Loss Function Weights and Hyperparameters Used in All Experiments.}
\label{tab:hyperparams}
\begin{tabular}{lcc}
\toprule
Hyperparameter & Symbol & Value \\
\midrule
Attribution loss weight & $\lambda_a$ & 0.3 \\
Contrastive loss weight & $\lambda_c$ & 0.1 \\
Fingerprint loss weight & $\lambda_f$ & 0.2 \\
Centroid loss weight    & $\lambda_r$ & 0.05 \\
Focal loss $\alpha$     & $\alpha$    & 0.75 \\
Focal loss $\gamma$     & $\gamma$    & 2.0 \\
Contrastive temperature & $\tau$      & 0.07 \\
Centroid momentum       & $m$         & 0.9 \\
Learning rate           & --          & $10^{-4}$ \\
Weight decay            & --          & $10^{-4}$ \\
Batch size              & --          & 4 \\
Epochs                  & --          & 50 \\
Gradient clip norm      & --          & 1.0 \\
\bottomrule
\end{tabular}
\end{table}

\subsection{Ablation Study}
\label{sec:ablation}

\begin{figure}[t]
    \centering
    \includegraphics[width=\columnwidth, trim=0 0 0 30, clip]{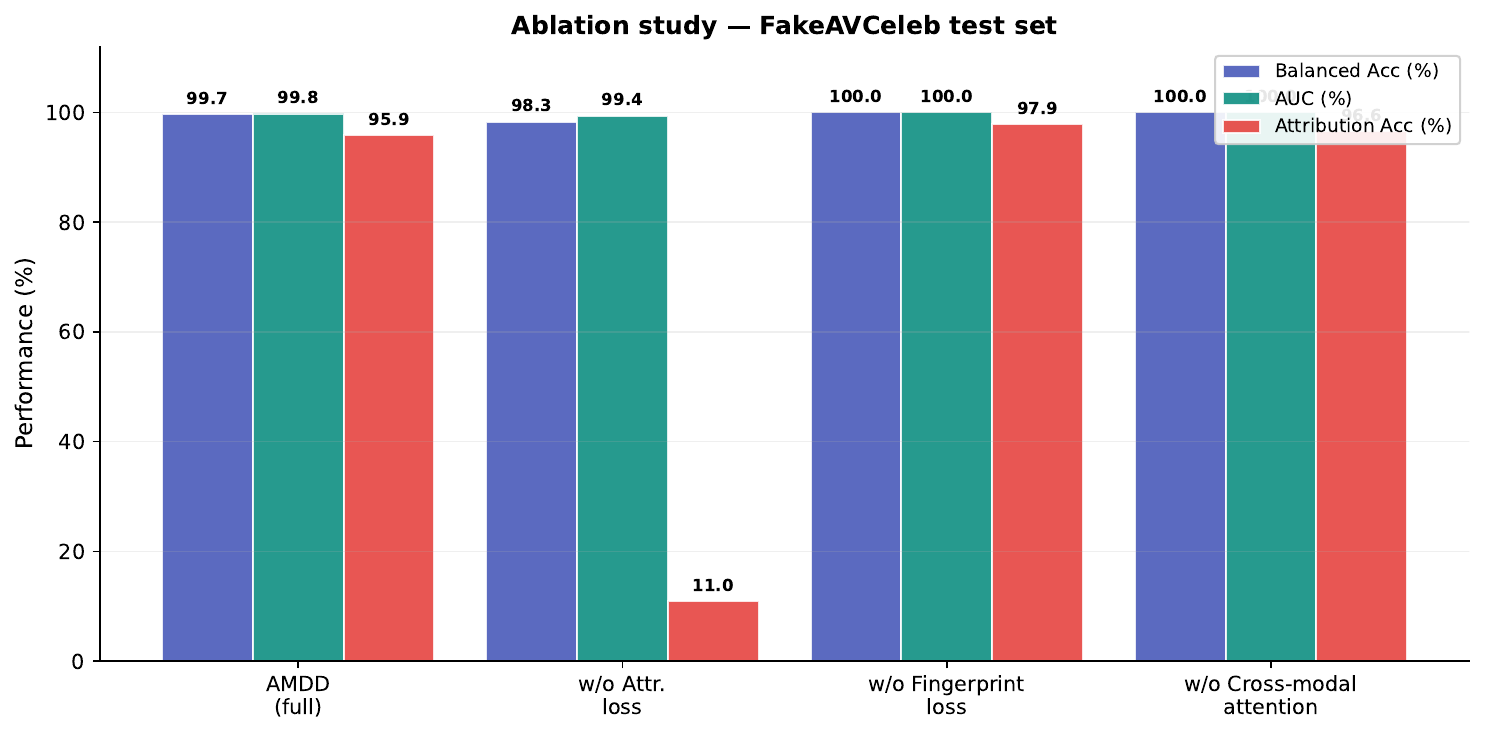}
    \caption{Ablation study comparing balanced accuracy, AUC, and attribution accuracy. Removing the attribution loss ($\mathcal{L}_{\text{attr}}$) collapses attribution accuracy from 95.9\% to 11.0\%, far below random chance, confirming its essential role in structured representation learning.}
    \label{fig:ablation}
\end{figure}

To validate each component of AMDD, we train three ablated variants by disabling individual loss terms while keeping all other components fixed. Table~\ref{tab:ablation} presents in-domain results on the FakeAVCeleb test set, and Fig.~\ref{fig:ablation} visualizes the comparison across all metrics.

\begin{table}[t]
\centering
\caption{Ablation Study on FakeAVCeleb Test Set. Removing Attribution Loss Collapses Generator Attribution to Below Random Chance (25\% for 4 Classes). All Metrics in \%.}
\label{tab:ablation}
\begin{tabular}{lcccccc}
\toprule
Model variant & Bal.~Acc & AUC & F1 & Real & Fake & Attr. \\
\midrule
\textbf{AMDD (full)}            & \textbf{99.7} & \textbf{99.8} & \textbf{99.7} & \textbf{100.0} & \textbf{99.3} & \textbf{95.9} \\
w/o $\mathcal{L}_{\text{attr}}$ & 98.3  & 99.4  & 99.0  & 97.3  & 99.3  & 11.0 \\
w/o $\mathcal{L}_{\text{fp}}$   & 100.0 & 100.0 & 100.0 & 100.0 & 100.0 & 97.9 \\
w/o $\mathcal{L}_{\text{cma}}$  & 100.0 & 100.0 & 100.0 & 100.0 & 100.0 & 96.6 \\
\bottomrule
\multicolumn{7}{l}{\footnotesize Bal.~Acc = balanced accuracy. Attr. = attribution accuracy.} \\
\multicolumn{7}{l}{\footnotesize $\mathcal{L}_{\text{fp}}$: fingerprint loss. $\mathcal{L}_{\text{cma}}$: cross-modal attention loss.}
\end{tabular}
\end{table}

The ablation reveals a critical finding: removing the attribution loss causes attribution accuracy to collapse from 95.9\% to 11.0\%, below random chance for four classes, confirming that attribution supervision is essential for learning generator identity. Crucially, binary detection accuracy remains high across all variants, which might superficially suggest that the attribution loss is redundant. It is not. A model that achieves near-perfect detection without attribution supervision is almost certainly exploiting a simpler, shallower signal, such as dataset-specific compression artifacts or background statistics, that happens to correlate with the real/fake label in this particular test set. The t-SNE analysis in Fig.~\ref{fig:tsne} makes this concrete: only the full model produces embedding clusters organized by generator identity, which is the structure one would expect from genuine forensic fingerprint learning. The marginal detection improvements seen in the w/o $\mathcal{L}_{\text{fp}}$ and w/o $\mathcal{L}_{\text{cma}}$ variants over the full model reflect threshold sensitivity on the small 220-sample test set rather than genuine performance gains; the AUC differences of at most 0.2\% are within statistical noise for this dataset size, and the attribution accuracy differences further confirm that these variants are not learning richer representations. The variants without attribution loss produce embeddings that separate real from fake but do not cluster by generator, indicating a less structured and less interpretable representation geometry.

\subsection{In-Domain Performance}

\begin{figure}[t]
    \centering
    \includegraphics[width=\columnwidth]{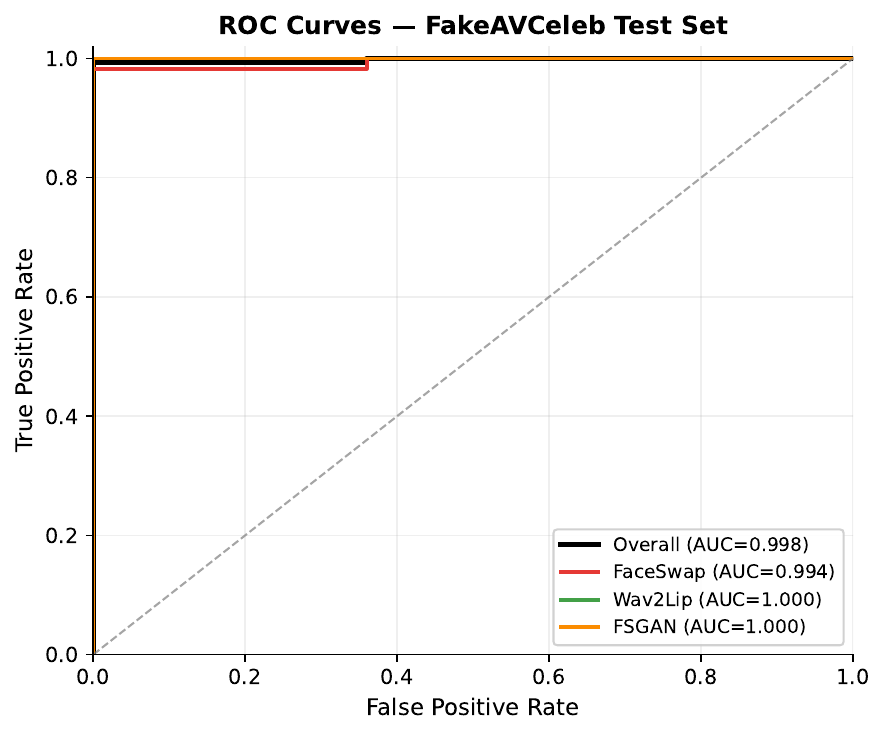}
    \caption{ROC curves on the FakeAVCeleb test set. The overall AUC of 99.8\% demonstrates strong discriminability. Wav2Lip and FSGAN achieve perfect AUC of 1.000; FaceSwap achieves 0.994.}
    \label{fig:roc}
\end{figure}

\begin{figure}[t]
    \centering
    \includegraphics[width=\columnwidth]{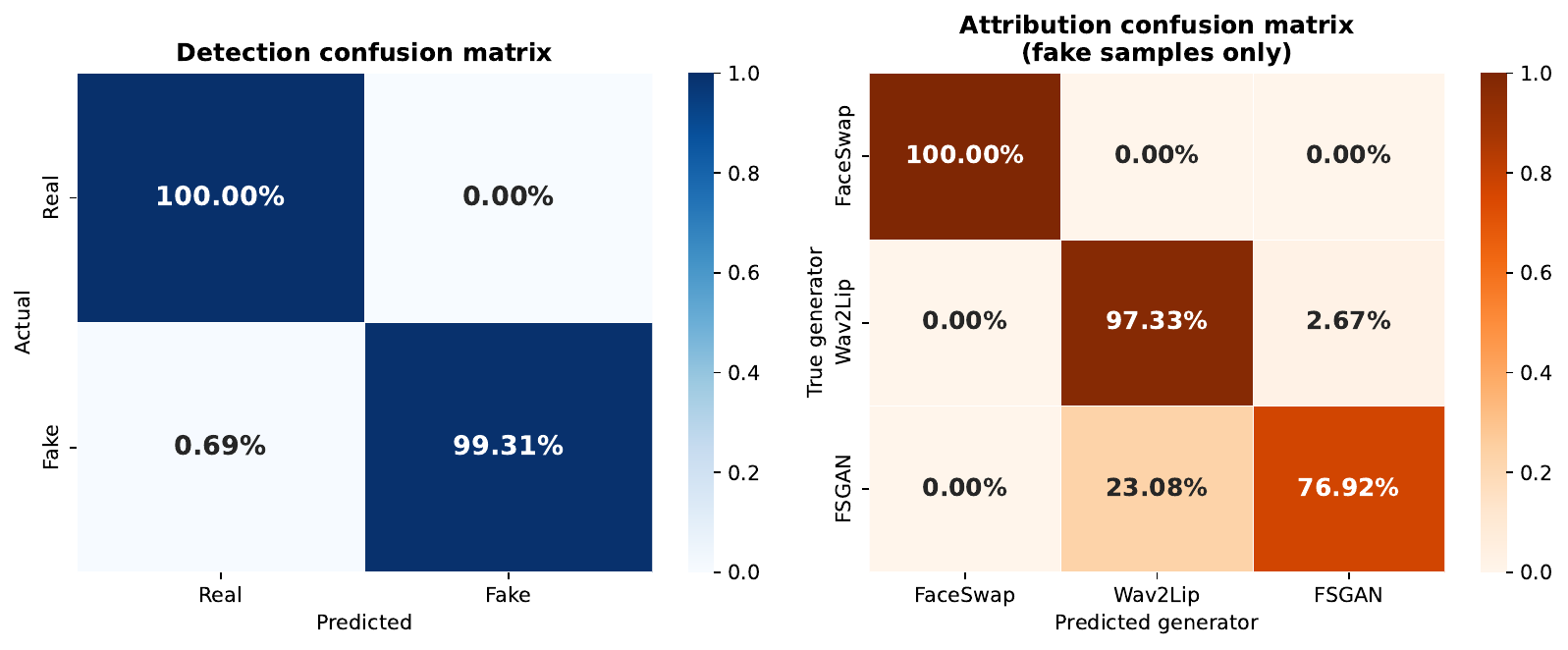}
    \caption{Confusion matrices for binary detection (left) and generator attribution (right). Detection confusion is minimal. Attribution confusion between FaceSwap and FSGAN reflects shared visual artifact patterns between these two face-swap methods.}
    \label{fig:confusion}
\end{figure}

\begin{figure*}[t]
    \centering
    \includegraphics[width=\textwidth]{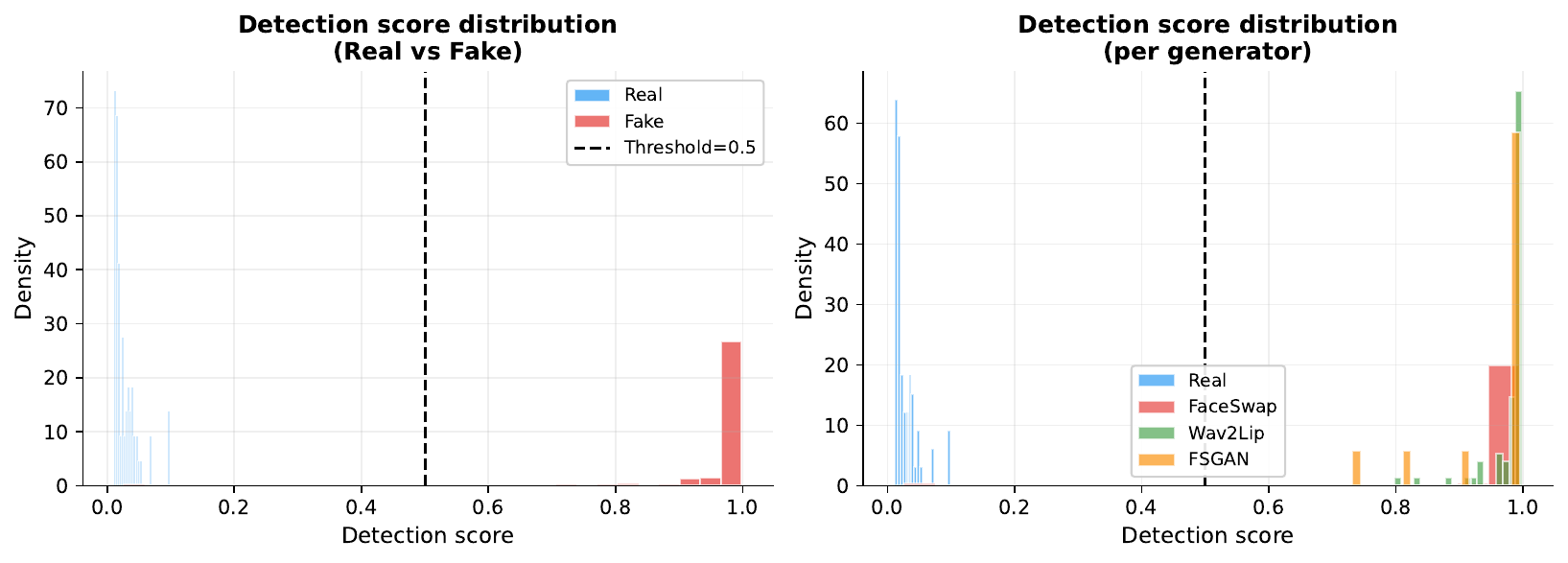}
    \caption{Detection score distributions on the FakeAVCeleb test set. Left: real vs.\ fake overall. Right: per-generator breakdown. Clear bimodal separation confirms well-calibrated detection scores across all generators.}
    \label{fig:scores}
\end{figure*}

Table~\ref{tab:per_gen} presents per-generator detection accuracy on the FakeAVCeleb test set. FaceSwap achieves 98.2\% detection accuracy, slightly below the perfect scores for Wav2Lip and FSGAN, which is consistent with the attribution confusion matrix in Fig.~\ref{fig:confusion}: some FaceSwap samples are misattributed to FSGAN, suggesting that these two face-swap methods share overlapping visual artifact characteristics, likely because both operate by warping and blending facial regions with similar boundary handling strategies. Wav2Lip, which modifies the lip region specifically to synchronize with target speech rather than performing full face replacement, produces more visually distinctive artifacts and is detected and attributed with perfect accuracy. Fig.~\ref{fig:roc} confirms consistently strong discriminative performance across all three generators, with per-generator AUC values ranging from 0.994 (FaceSwap) to 1.000 (Wav2Lip and FSGAN). Fig.~\ref{fig:scores} shows that detection score distributions are strongly bimodal: real samples cluster near zero and fake samples cluster near one, with minimal overlap, indicating that the model produces well-calibrated confidence estimates rather than marginal decisions near the threshold.

\begin{table}[t]
\centering
\caption{Per-Generator Detection and Attribution Accuracy on the FakeAVCeleb Test Set. $^\ddagger$Result Based on 13 Samples; Interpret with Caution.}
\label{tab:per_gen}
\begin{tabular}{lccc}
\toprule
Generator & Test Samples & Detection Acc (\%) & Attribution Acc (\%) \\
\midrule
FaceSwap          & 57  & 98.2  & \multirow{3}{*}{95.9 (overall)} \\
Wav2Lip           & 75  & 100.0 & \\
FSGAN$^\ddagger$  & 13  & 100.0 & \\
\midrule
Real      & 75  & 100.0 & -- \\
\midrule
\textbf{Overall} & \textbf{220} & \textbf{99.7} & \textbf{95.9} \\
\bottomrule
\end{tabular}
\end{table}

\subsection{Comparison with State-of-the-Art Methods}

Table~\ref{tab:sota} compares AMDD against state-of-the-art methods on FakeAVCeleb. AMDD achieves 99.8\% AUC and 99.7\% balanced accuracy, competitive with the strongest prior methods while operating without access to pre-trained large-scale audio-visual models. The comparison should be interpreted carefully: CMALDD-PTAF~\cite{cmaldd} is evaluated on a different subset of FakeAVCeleb than AMDD, making direct numerical comparison imprecise, and several methods in the table do not report balanced accuracy, making head-to-head comparison on that metric impossible. What is unambiguous is that AMDD is the only framework in this comparison that provides generator attribution at 95.9\% accuracy alongside detection, a capability with practical value for forensic investigation and accountability that binary detectors cannot offer regardless of their detection score. The attribution head adds no inference-time cost beyond a small MLP applied to the already-computed fused embedding, making this a free capability from a deployment perspective.

\begin{table}[t]
\centering
\caption{Comparison with State-of-the-Art Methods on FakeAVCeleb. V = visual only, A+V = audio-visual. $^\dagger$Evaluated on different subsets or protocols. -- = not reported. AMDD is the only method providing attribution.}
\label{tab:sota}
\begin{tabular}{lcccc}
\toprule
Method & Mod. & AUC (\%) & Bal.~Acc (\%) & Attr. (\%) \\
\midrule
XceptionNet~\cite{xception}         & V   & 88.1  & --    & -- \\
LipForensics~\cite{lipforensics}    & V   & 93.4  & --    & -- \\
ResViT~\cite{resvit}                & V   & --    & 88.1  & -- \\
MDS~\cite{mds}                      & A+V & --    & --    & -- \\
Joint AV~\cite{jointAV}             & A+V & 92.3  & --    & -- \\
NPVForensics~\cite{npvforensics}    & A+V & 99.0  & --    & -- \\
CMALDD-PTAF~\cite{cmaldd}$^\dagger$ & A+V & 99.6  & --    & -- \\
\midrule
\textbf{AMDD (ours)} & A+V & \textbf{99.8} & \textbf{99.7} & \textbf{95.9} \\
\bottomrule
\end{tabular}
\end{table}

\subsection{Cross-Dataset Generalization}

\begin{figure}[t]
    \centering
    \includegraphics[width=\columnwidth]{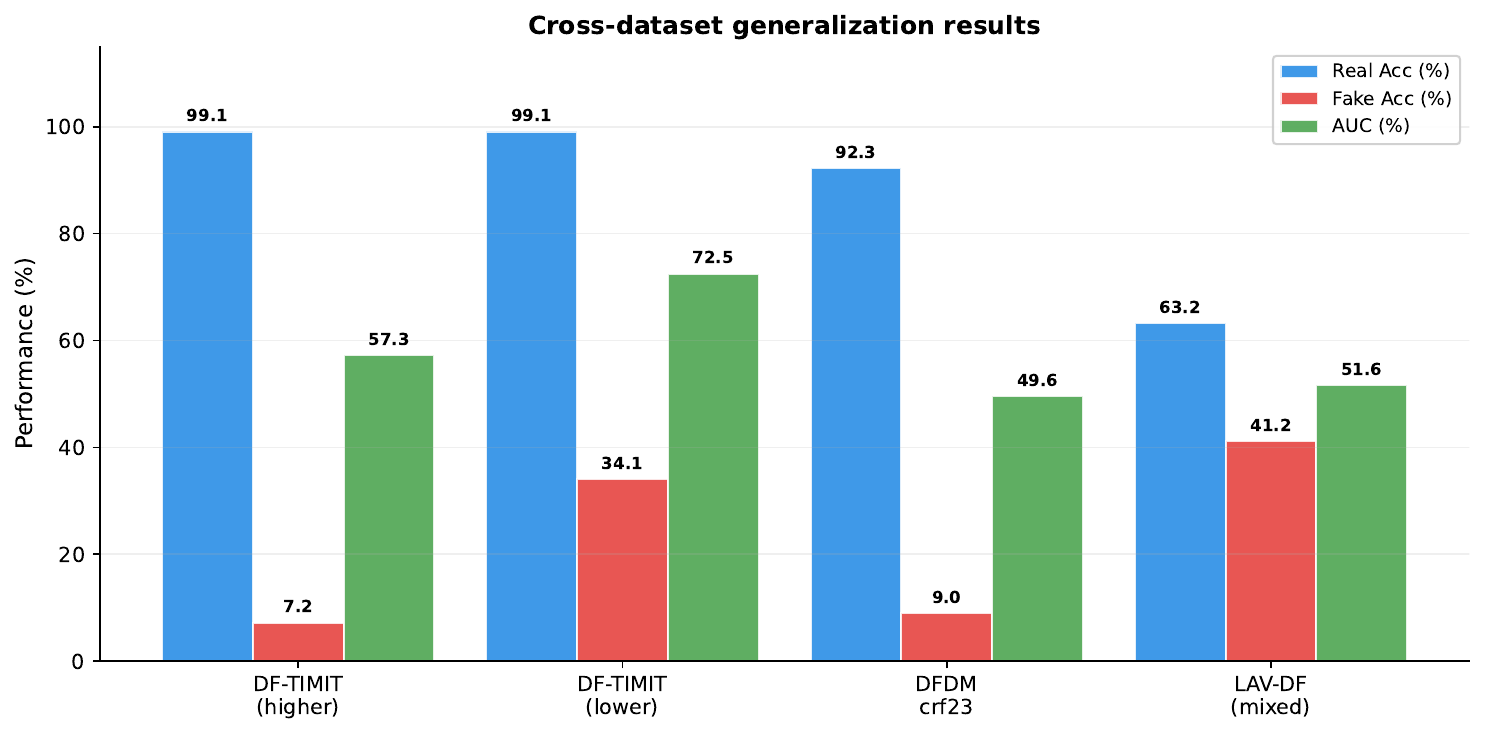}
    \caption{Cross-dataset generalization results. Real detection accuracy (blue) remains consistently high across datasets while fake detection (red) varies significantly with generator distribution shift. AUC (green) provides threshold-independent discriminability.}
    \label{fig:crossdataset}
\end{figure}

Table~\ref{tab:cross} presents cross-dataset evaluation results. Fig.~\ref{fig:crossdataset} visualizes the key trends across datasets and conditions.

\begin{table*}[t]
\centering
\caption{Cross-Dataset Generalization Results. All Generators Are Unseen During Training. DFDM Contains No Audio; Silent Audio Is Used as Fallback. $^\dagger$DFDM crf10 Threshold Collapses Due to Distribution Shift; AUC (Threshold-Free) Reflects True Discriminability.}
\label{tab:cross}
\begin{tabular}{llccccc}
\toprule
Dataset & Condition & Bal.~Acc (\%) & AUC (\%) & F1 (\%) & Real Acc (\%) & Fake Acc (\%) \\
\midrule
\multirow{2}{*}{DF-TIMIT~\cite{dftimit}}
  & Higher quality     & 53.1 & 57.3 & 13.3 & 99.1 & 7.2  \\
  & Lower quality      & 66.6 & 72.5 & 50.5 & 99.1 & 34.1 \\
\midrule
\multirow{3}{*}{DFDM~\cite{dfdm}}
  & crf0 (lossless)        & 50.3 & 49.6 & 2.7  & 99.2 & 1.3  \\
  & crf10 (high)$^\dagger$ & 50.2 & 49.6 & 88.8 & 0.8  & 99.6 \\
  & crf23 (low)            & 50.7 & 49.6 & 16.3 & 92.3 & 9.0  \\
\midrule
\multirow{3}{*}{LAV-DF~\cite{lavdf}}
  & Video-only fake    & \multirow{3}{*}{52.2} & \multirow{3}{*}{51.6} & \multirow{3}{*}{46.2} & \multirow{3}{*}{63.2} & 42.8 \\
  & Audio-only fake    & & & & & 42.2 \\
  & Both manipulated   & & & & & 38.6 \\
\bottomrule
\end{tabular}
\end{table*}

\textbf{Real video detection generalizes robustly.} Across all datasets, real video detection accuracy stays above 92\% in all but one anomalous threshold-collapse case, suggesting the model picks up on properties of authentic video that transcend generator identity, likely related to natural temporal statistics and audio-visual coherence that no current forgery pipeline fully replicates.

\textbf{Fake detection does not transfer.} Detection of forged content from unseen generators is consistently poor, confirming the widely reported limitation of fingerprint-based approaches: a model trained on Wav2Lip, FaceSwap, and FSGAN has learned those generators' specific traces, not a general theory of forgery. This is not a specific AMDD failure; it is a property of the problem, and methods claiming otherwise typically benefit from data set contamination or overlapping generator families. This behavior is consistent with domain adaptation theory~\cite{domainadapt}, which establishes that classifier performance degrades in proportion to the divergence between training and test distributions. In the deepfake context, each generator defines its own artifact distribution, and cross-generator transfer requires either substantial overlap between those distributions or explicit domain-invariant feature learning, neither of which holds when generators differ fundamentally in their blending pipelines.

\textbf{Compression and audio availability matter.} Lower-quality video exposes more visible artifacts and is easier to detect across datasets. DFDM, the only audio-free dataset, produces the weakest results, confirming that forcing a multimodal model to process silent audio breaks the cross-modal fusion on which it depends.

\begin{figure*}[!t]
    \centering
    \includegraphics[width=\textwidth]{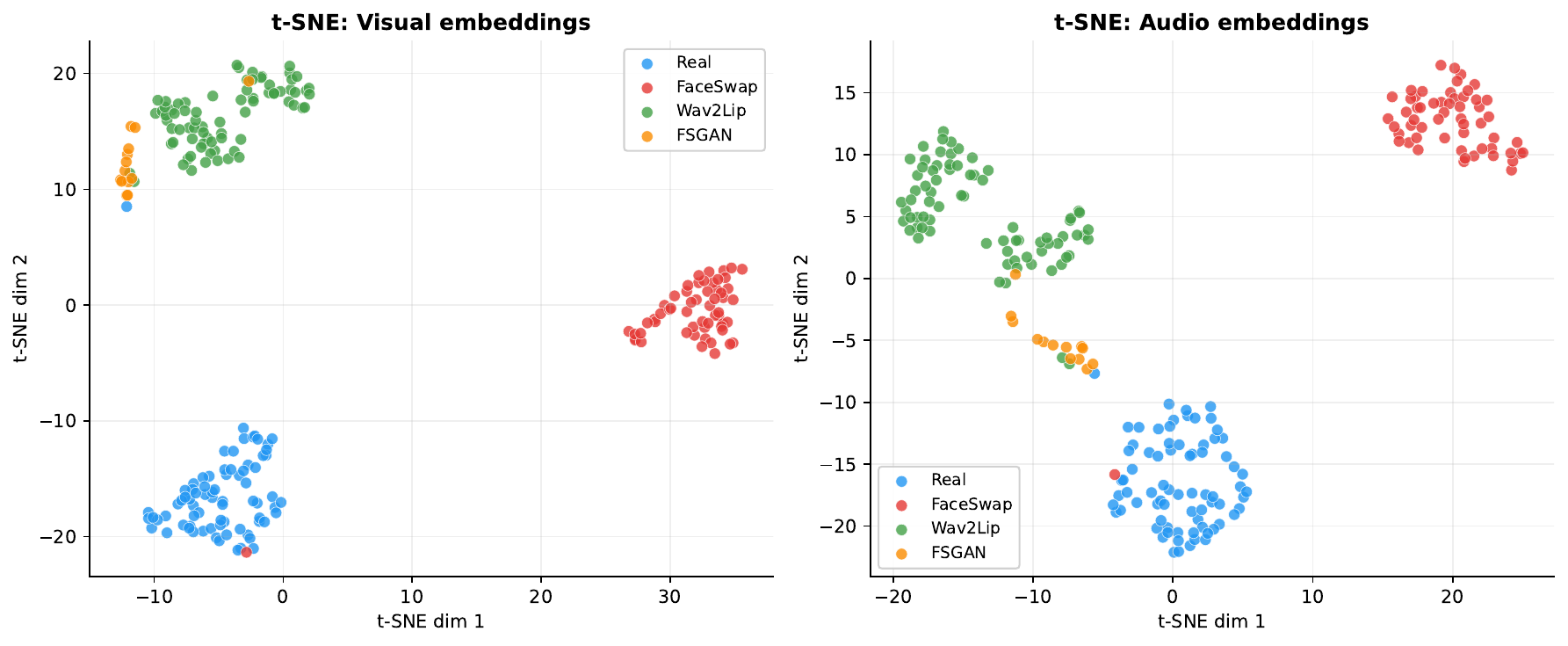}
    \caption{t-SNE visualization of unimodal embeddings before fusion. Visual embeddings (left) show stronger generator clustering than audio embeddings (right), confirming that visual features carry richer generator-specific fingerprints for the FakeAVCeleb generators.}
    \label{fig:tsne_modalities}
\end{figure*}
% ==================================================
\section{Discussion}
\label{sec:discussion}

\subsection{What the Embedding Space Reveals}

\begin{figure*}[!t]
    \centering
    \includegraphics[width=\textwidth]{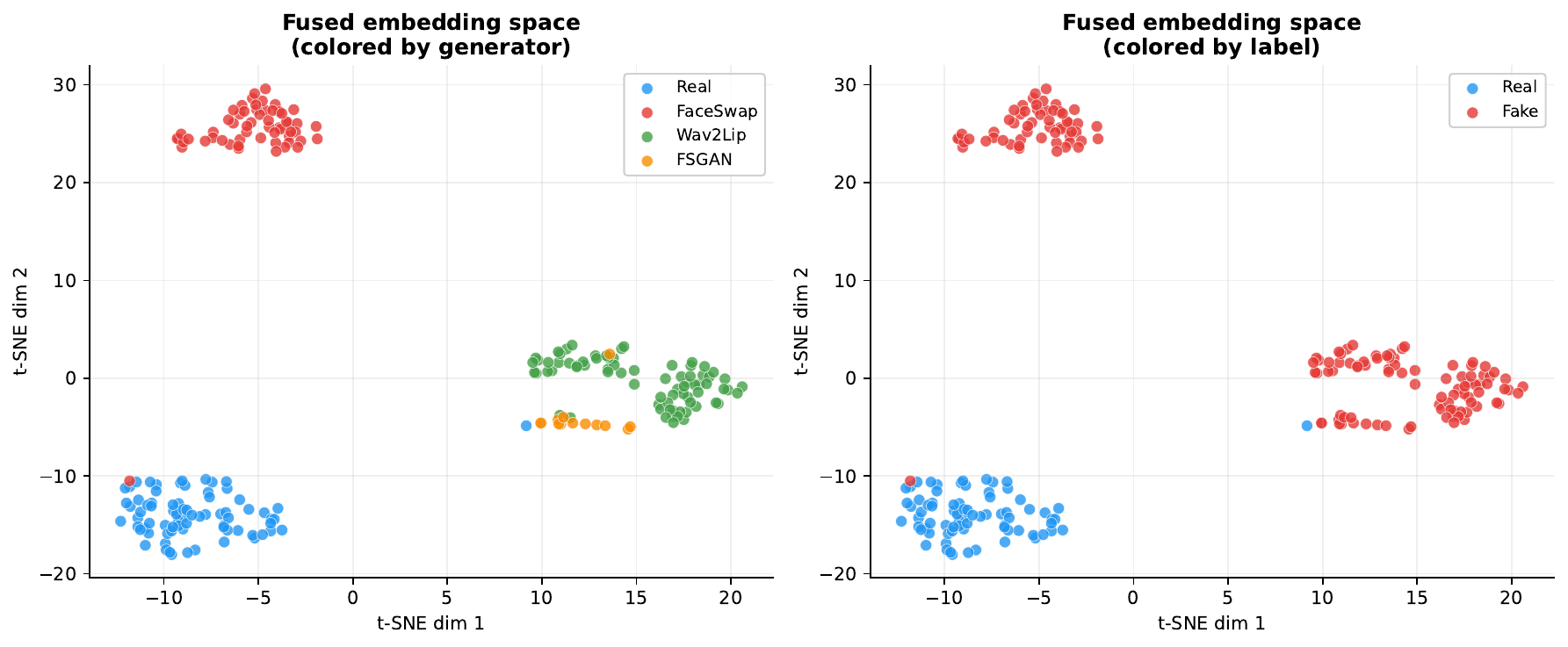}
    \caption{t-SNE visualization of fused embeddings on the FakeAVCeleb test set. Left: colored by generator identity, showing distinct generator-specific clusters. Right: colored by binary label, showing clear real/fake separation. The structured geometry confirms that attribution-guided learning captures generator-specific forensic fingerprints.}
    \label{fig:tsne}
\end{figure*}

The t-SNE projections in Fig.~\ref{fig:tsne} are perhaps the clearest evidence that attribution guidance changes what the model learns. Without attribution, a detector optimized purely for binary classification has no incentive to separate Wav2Lip embeddings from FaceSwap embeddings. Both are ``fake'', and conflating them costs nothing in terms of the training loss. With attribution, the model must learn features that are discriminative across generators, which forces the representation toward the forensically meaningful structure visible in the left panel of Fig.~\ref{fig:tsne}: distinct generator clusters that also happen to separate cleanly by label.

Comparing visual and audio embeddings before fusion (Fig.~\ref{fig:tsne_modalities}) is equally informative: generator clusters are much more pronounced in the visual stream than in the audio stream, consistent with the fact that FakeAVCeleb's fake samples manipulate video while retaining real audio. The audio encoder is learning to discriminate, but it has less signal to work with in this dataset.

\subsection{Cross-Modal Alignment}

\begin{figure}[t]
    \centering
    \includegraphics[width=\columnwidth, trim=0 0 0 20, clip]{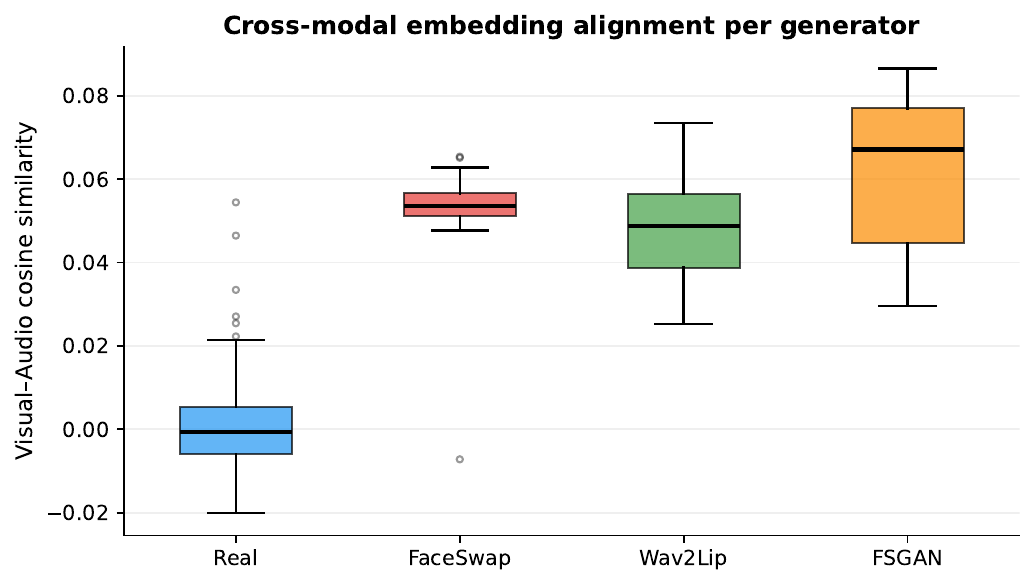}
    \caption{Cross-modal visual--audio cosine similarity per generator. Real samples show higher alignment reflecting natural audio-visual coherence. Generator-specific patterns confirm distinct cross-modal artifact profiles induced by different generators.}
    \label{fig:crossmodal}
\end{figure}

Fig.~\ref{fig:crossmodal} shows that real samples consistently achieve higher visual-audio cosine similarity than fake samples, with generator-specific variation in how much misalignment is introduced. This is the signal that the CMFFC loss is designed to amplify: by pulling same-generator visual and audio projections toward alignment, the loss encourages the encoder to treat cross-modal coherence as a forensic feature rather than noise. The fact that generator-specific patterns are visible in this plot, without any explicit per-generator supervision on similarity, suggests the fingerprint structure is genuinely there to be learned.

\subsection{Where the Model Looks: GradCAM Analysis}

\begin{figure}[!t]
    \centering
    \includegraphics[width=\columnwidth, trim=0 0 0 120, clip]{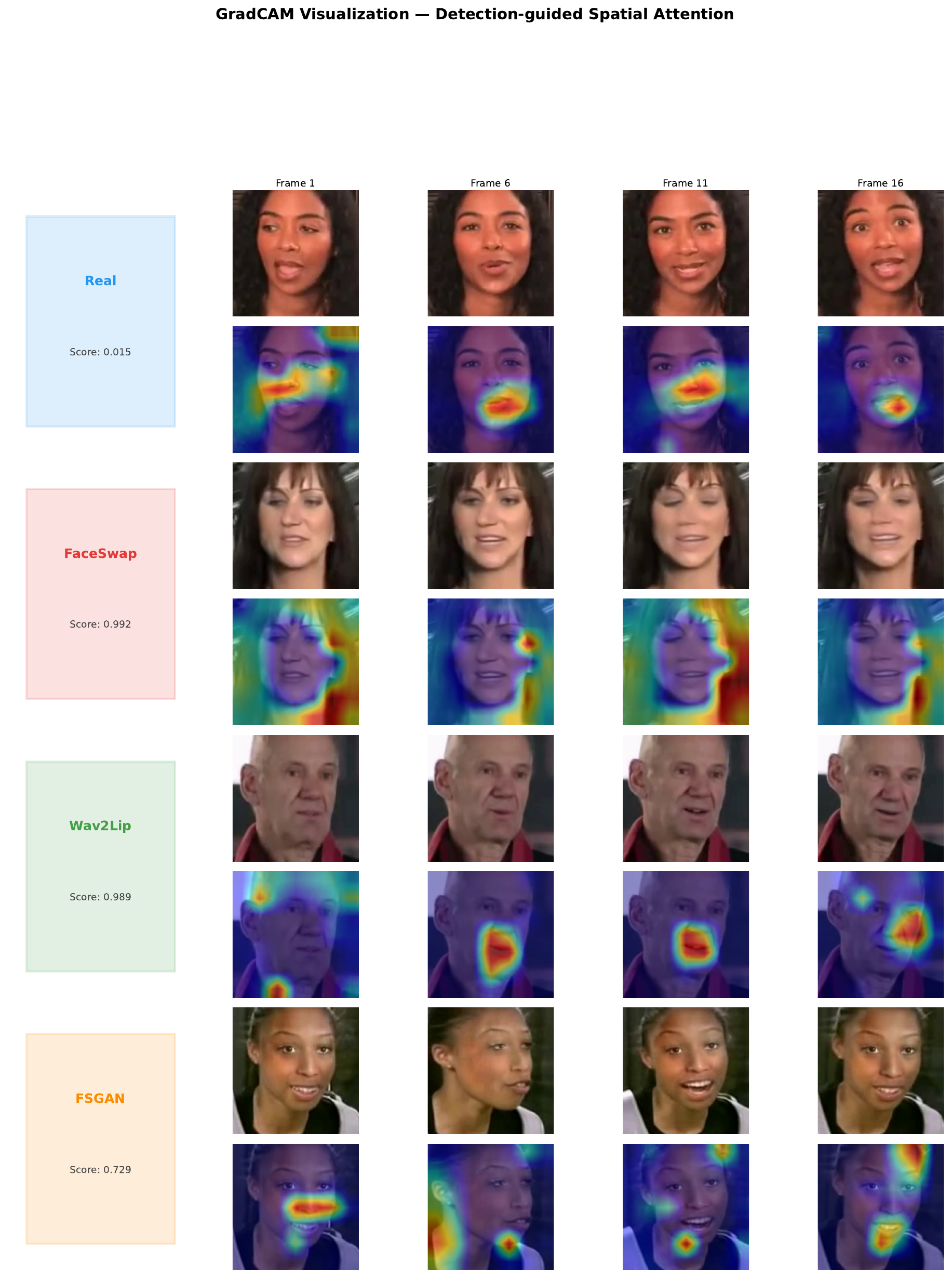}
    \caption{GradCAM visualization of detection-guided spatial attention for real and three fake generator samples. The model consistently focuses on facial boundary regions (mouth, eyes, and chin) where face-swap artifacts are most prominent. Higher activation in fake samples reflects stronger manipulation traces.}
    \label{fig:gradcam}
\end{figure}

The GradCAM maps in Fig.~\ref{fig:gradcam} show that the model's attention concentrates on the facial boundary regions (mouth corners, eye contours, and chin) rather than on background or hair regions that would indicate the model is exploiting identity cues. This is reassuring: these are precisely the regions where blending artifacts from face-swap methods are most commonly found. Compared to the near-uniform low activation on the real sample, the fake samples show concentrated high-activation patches that vary in location by generator, consistent with the different blending strategies used by FaceSwap, Wav2Lip, and FSGAN.

\subsection{Mel Spectrogram Analysis}

\begin{figure*}[!t]
    \centering
    \includegraphics[width=\textwidth, trim= 0 18 0 20, clip]{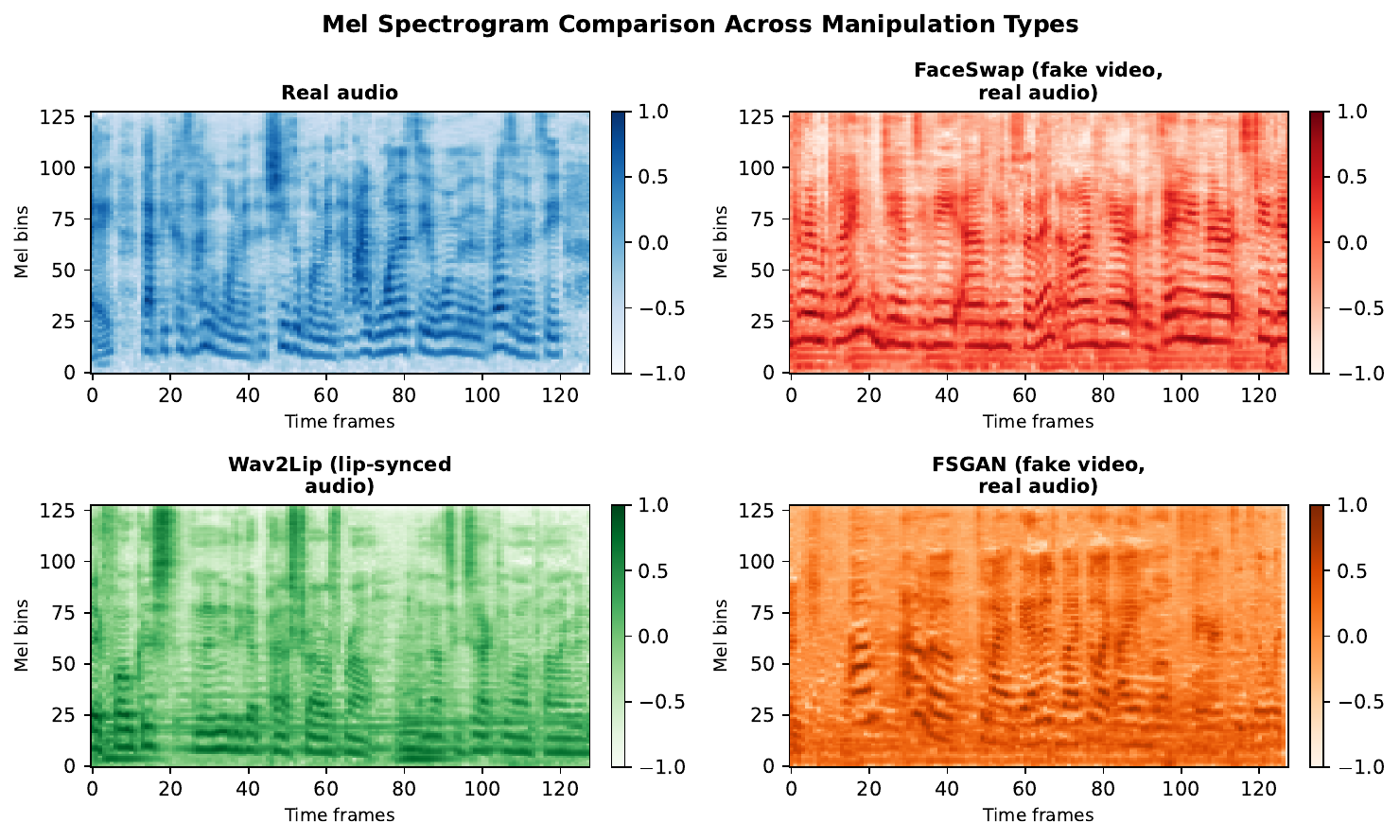}
    \caption{Mel spectrogram comparison across manipulation types. Real audio exhibits natural harmonic patterns. FaceSwap and FSGAN retain real audio characteristics since only the visual modality is manipulated. This validates the use of FakeVideo-RealAudio samples for training.}
    \label{fig:mel}
\end{figure*}

Fig.~\ref{fig:mel} confirms that the FakeVideo-RealAudio design choice is acoustically sound: the mel spectrograms of FaceSwap and FSGAN samples are visually identical to real audio, as expected when audio tracks are untouched. This means the audio encoder has no spectral signal to exploit for detection in these cases and must rely on what the cross-modal attention module can infer from the visual-audio relationship. That the model still achieves strong detection confirms that the visual stream carries sufficient information, and the audio stream's value lies in providing a consistency reference rather than an independent forgery signal.

\subsection{Attribution as Regularization: What the Ablation Tells Us}

The most striking finding in the ablation (Table~\ref{tab:ablation}) is not that detection degrades when attribution is removed, as it largely does not, but that attribution collapses to 11\% accuracy, far below random chance for four classes. This asymmetry tells a clear story: binary detection is an easier problem that can be solved with a shallower, less structured representation, while attribution requires the model to learn features that discriminate between forgery types. Removing the attribution loss allows the model to take this easier path. The value of attribution-guided training, then, is not in the numbers on the in-domain detection table but in the structure of the learned representation, a structure that the t-SNE visualizations and cross-modal alignment plots make tangible.

To understand why attribution acts as regularization rather than merely adding a task, it helps to think about the geometry of the embedding space under each objective. A binary detection loss only requires that real and fake embeddings be linearly separable, which is a weak geometric constraint achievable by many different representations. An attribution loss additionally requires that embeddings from different generators be separable from each other, which is a strictly stronger constraint that can only be satisfied by representations that encode generator-specific information. The detection objective is subsumed by the attribution objective in the sense that any embedding space that separates generators will also separate real from fake, but the converse is not true. Attribution supervision therefore forces the model toward a higher-fidelity representation of the input than detection alone would require, and this improved fidelity is what we observe in the t-SNE clusters and cross-modal alignment patterns.

\subsection{Generalization Limits and What They Mean}

The cross-dataset results are honest: the model does not generalize well to unseen generators. We interpret this not as a failure of AMDD specifically but as a consequence of the forensic fingerprint paradigm. A model trained to recognize three generator families will learn those families' signatures. When tested on DFaker, DFL, or IAE, generators that use entirely different blending pipelines, the learned signatures do not transfer. Improving generalization likely requires either training on a much wider distribution of generators, designing generator-agnostic anomaly features, or both. Unimodal fallback handling and cross-modal imputation for audio-absent datasets represent more tractable near-term improvements.

\subsection{Limitations}

The small test set for FSGAN (13 samples) makes per-generator attribution statistics unreliable for that class specifically. The model's dependence on face detection for preprocessing creates a fragility that degrades gracefully but not gracefully enough for in-the-wild deployment. Training was conducted on a single dataset family; the generalization behavior documented here is the realistic expectation for this setting, not an anomaly. Finally, the DFDM evaluation is handicapped by missing audio, a limitation of the dataset rather than the method, but one that a production system would need to handle explicitly.

\subsection{Future Directions}

Three directions seem most productive. First, open-set attribution, that is detecting that a sample comes from a \textit{novel} generator rather than misclassifying it as one of the known ones, would substantially improve practical reliability. Second, domain-adaptive fine-tuning with limited unlabeled data from a new distribution could bridge the generalization gap without requiring full retraining. Third, explicit cross-modal imputation, where the model estimates a plausible audio embedding when audio is absent, would restore multimodal fusion capability on video-only datasets.

% ==================================================
\section{Conclusion}
\label{sec:conclusion}

We have presented AMDD, a multimodal deepfake detection framework built on the premise that a detector which can identify the source generator is learning something more forensically meaningful than one which cannot. By jointly training detection and attribution within a shared embedding space and enforcing cross-modal fingerprint consistency, we obtain representations that cluster by generator identity, attend to semantically relevant facial regions, and maintain strong audio-visual alignment on authentic content. On FakeAVCeleb, the framework achieves 99.7\% balanced accuracy, 99.8\% AUC, and 95.9\% attribution accuracy. The ablation establishes that attribution supervision is load-bearing: removing it collapses attribution accuracy from 95.9\% to 11\% while leaving detection superficially intact, which is precisely the scenario where a model has learned the wrong thing. Cross-dataset evaluation on three held-out datasets confirms that real video detection generalizes well, while fake detection on unseen generators does not, a result we report without hedging, as it reflects a genuine property of fingerprint-based forensics rather than a limitation of our implementation. We hope this work is useful both as a practical framework and as an honest characterization of where the field currently stands.

% ==================================================
\section*{Acknowledgment}

This work was supported by the National Natural Science Foundation of China under Grant 92152109.

\bibliographystyle{IEEEtran}
\bibliography{references}

@article{capst2025,
  title={CapST: Leveraging Capsule Networks and Temporal Attention for Accurate Model Attribution in Deepfake Videos},
  author={Ahmad, Wasim and Peng, Yan-Tsung and Chang, Yuan-Hao and Ganfure, Gaddisa Olani and Khan, Sarwar},
  journal={ACM Transactions on Multimedia Computing, Communications and Applications},
  volume={21},
  number={4},
  pages={1--23},
  year={2025}
}

@article{fame2025,
  title={{FAME}: A Lightweight Spatio-Temporal Network for Model Attribution of Face-Swap Deepfakes},
  author={Ahmad, Wasim and Peng, Yan-Tsung and Chang, Yuan-Hao},
  journal={Expert Systems with Applications},
  volume={292},
  pages={128571},
  year={2025}
}

@article{resvit,
  title={ResViT: A Framework for Deepfake Videos Detection},
  author={Ahmad, Wasim and Ali, Imad and Shahzad, Adil and Hashmi, Ammarah and Ghaffar, Faisal},
  journal={International Journal of Electrical and Computer Engineering Systems},
  volume={13},
  number={9},
  pages={807--813},
  year={2022},
  doi={10.32985/ijeces.13.9.9}
}

@inproceedings{hashmi2022,
  title={Multimodal Forgery Detection Using Ensemble Learning},
  author={Hashmi, Ammarah and Shahzad, Sahibzada Adil and Ahmad, Wasim and Lin, Chia-Wen and Tsao, Yu and Wang, Hsin-Min},
  booktitle={2022 Asia-Pacific Signal and Information Processing Association Annual Summit and Conference (APSIPA ASC)},
  pages={1524--1532},
  year={2022}
}

@misc{faceswap,
  title={Faceswap},
  author={deepfakes},
  howpublished={\url{https://github.com/deepfakes/faceswap}},
  year={2017}
}

@inproceedings{fsgan,
  title={{FSGAN}: Subject Agnostic Face Swapping and Reenactment},
  author={Nirkin, Yuval and Keller, Yosi and Hassner, Tal},
  booktitle={Proceedings of the IEEE/CVF International Conference on Computer Vision (ICCV)},
  pages={7184--7193},
  year={2019}
}

@inproceedings{wav2lip,
  title={A Lip Sync Expert Is All You Need for Speech to Lip Generation In The Wild},
  author={Prajwal, K R and Mukhopadhyay, Rudrabha and Namboodiri, Vinay P and Jawahar, C V},
  booktitle={Proceedings of the 28th ACM International Conference on Multimedia},
  pages={484--492},
  year={2020}
}

@inproceedings{sv2tts,
  title={Transfer Learning from Speaker Verification to Multispeaker Text-To-Speech Synthesis},
  author={Jia, Ye and Zhang, Yu and Weiss, Ron J and Wang, Quan and Shen, Jonathan and Ren, Fei and Nguyen, Patrick and Pang, Ruoming and Moreno, Ignacio Lopez and Wu, Yonghui and others},
  booktitle={Advances in Neural Information Processing Systems (NeurIPS)},
  volume={31},
  year={2018}
}

@inproceedings{fakeavceleb,
  title={{FakeAVCeleb}: A Novel Audio-Video Multimodal Deepfake Dataset},
  author={Khalid, Hasam and Tariq, Shahroz and Kim, Minha and Woo, Simon S},
  booktitle={Advances in Neural Information Processing Systems (NeurIPS) Datasets and Benchmarks Track},
  year={2021}
}

@inproceedings{lavdf,
  title={Do You Really Mean That? Content Driven Audio-Visual Deepfake Dataset and Multimodal Method for Temporal Forgery Localization},
  author={Cai, Zhixi and Stefanov, Kalin and Dhall, Abhinav and Hayat, Munawar},
  booktitle={2023 IEEE International Conference on Digital Image Computing: Techniques and Applications (DICTA)},
  year={2022}
}

@inproceedings{dftimit,
  title={Deepfakes and Beyond: A Survey of Face Manipulation and Fake Detection},
  author={Tolosana, Ruben and Vera-Rodriguez, Ruben and Fierrez, Julian and Morales, Aythami and Ortega-Garcia, Javier},
  booktitle={Information Fusion},
  volume={64},
  pages={131--148},
  year={2020}
}

@article{dfdm,
  title={Model Attribution of Face-swap Deepfake Videos},
  author={Jia, Shan and Li, Xin and Lyu, Siwei},
  journal={arXiv preprint arXiv:2202.12951},
  year={2022}
}

@inproceedings{xception,
  title={Xception: Deep Learning with Depthwise Separable Convolutions},
  author={Chollet, Fran{\c{c}}ois},
  booktitle={Proceedings of the IEEE Conference on Computer Vision and Pattern Recognition (CVPR)},
  pages={1251--1258},
  year={2017}
}

@inproceedings{lipforensics,
  title={{LipForensics}: Breaking Temporal-Inconsistency Based DeepFake Detection},
  author={Haliassos, Alexandros and Vougioukas, Konstantinos and Petridis, Stavros and Pantic, Maja},
  booktitle={Proceedings of the IEEE/CVF Conference on Computer Vision and Pattern Recognition (CVPR)},
  pages={1795--1805},
  year={2021}
}

@inproceedings{recce,
  title={End-to-End Reconstruction-Classification Learning for Face Forgery Detection},
  author={Cao, Junyi and Ma, Chao and Yao, Taiping and Chen, Shen and Ding, Shouhong and Yang, Xiaokang},
  booktitle={Proceedings of the IEEE/CVF Conference on Computer Vision and Pattern Recognition (CVPR)},
  pages={4523--4533},
  year={2022}
}

@inproceedings{eyeblinking,
  title={In Ictu Oculi: Exposing AI Created Fake Videos by Detecting Eye Blinking},
  author={Li, Yuezun and Chang, Ming-Ching and Lyu, Siwei},
  booktitle={IEEE International Workshop on Information Forensics and Security (WIFS)},
  pages={1--7},
  year={2018}
}

@inproceedings{headpose,
  title={Exposing DeepFake Videos By Detecting Face Warping Artifacts},
  author={Li, Yuezun and Lyu, Siwei},
  booktitle={Proceedings of the IEEE/CVF Conference on Computer Vision and Pattern Recognition Workshops},
  year={2019}
}

@inproceedings{frequency,
  title={Leveraging Frequency Analysis for Deep Fake Image Recognition},
  author={Frank, Joel and Eisenhofer, Thorsten and Sch{\"o}nherr, Lea and Fischer, Asja and Kolossa, Dorothea and Holz, Thorsten},
  booktitle={International Conference on Machine Learning (ICML)},
  pages={3247--3258},
  year={2020}
}

@inproceedings{lstm,
  title={Exploiting Visual Artifacts to Expose Deepfakes with Face Warping Artifacts},
  author={Li, Yuezun and Lyu, Siwei},
  booktitle={IEEE Winter Applications of Computer Vision Workshops (WACVW)},
  year={2019}
}

@inproceedings{3dcnn,
  title={Detecting Deepfake Videos Using Euler Video Magnification},
  author={Qi, Hanxiang and Guo, Shengchao and Torchiani, Felix and Weuster, Daniel and Steinebach, Martin},
  booktitle={Media Watermarking, Security, and Forensics},
  year={2020}
}

@inproceedings{asvspoof,
  title={{ASVspoof} 2019: A Large-Scale Public Database of Synthesized, Converted and Replayed Speech},
  author={Wang, Xin and Yamagishi, Junichi and Todisco, Massimiliano and others},
  booktitle={Computer Speech \& Language},
  volume={64},
  pages={101114},
  year={2020}
}

@inproceedings{rawnet,
  title={{RawNet2}: Towards a More Interpretable Representation for Audio Spoofing Detection},
  author={Tak, Hemlata and Patino, Jose and Todisco, Massimiliano and Nautsch, Andreas and Evans, Nicholas and Larcher, Anthony},
  booktitle={Interspeech},
  pages={2586--2590},
  year={2021}
}

@inproceedings{aasist,
  title={{AASIST}: Audio Anti-Spoofing Using Integrated Spectro-Temporal Graph Attention Networks},
  author={Jung, Jee-weon and Heo, Hee-Soo and Tak, Hemlata and Shim, Hye-jin and Chung, Joon Son and Lee, Bong-Jin and Yu, Ha-Jin and Evans, Nicholas},
  booktitle={IEEE International Conference on Acoustics, Speech and Signal Processing (ICASSP)},
  pages={6367--6371},
  year={2022}
}

@inproceedings{mds,
  title={Detecting Deep-Fake Videos from Appearance and Behavior},
  author={Chugh, Komal and Gupta, Parul and Dhall, Abhinav and Subramanian, Ramanathan},
  booktitle={IEEE Winter Conference on Applications of Computer Vision (WACV)},
  year={2020}
}

@inproceedings{emotions,
  title={Emotions Don't Lie: An Audio-Visual Deepfake Detection Method Using Affective Cues},
  author={Mittal, Trisha and Bhattacharya, Uttaran and Chandra, Rohan and Bera, Aniket and Manocha, Dinesh},
  booktitle={Proceedings of the 28th ACM International Conference on Multimedia},
  pages={2823--2832},
  year={2020}
}

@inproceedings{jointAV,
  title={Joint Audio-Visual Deepfake Detection},
  author={Zhou, Yipin and Lim, Ser-Nam},
  booktitle={Proceedings of the IEEE/CVF International Conference on Computer Vision (ICCV)},
  pages={14800--14809},
  year={2021}
}

@inproceedings{cmaldd,
  title={Leveraging Large Language Models for Multimodal Deepfake Detection},
  author={Liu, Zhiyuan and Wang, Zhixin and Wen, Bihan},
  booktitle={Proceedings of the IEEE/CVF Conference on Computer Vision and Pattern Recognition (CVPR)},
  year={2024}
}

@inproceedings{npvforensics,
  title={{NPVForensics}: Jointing Non-critical Phonemes and Visemes for Deepfake Detection},
  author={Feng, Hao and Zhou, Jikang and Han, Jianguo and Ni, Rongrong and Zhao, Yao and He, Chao},
  booktitle={Proceedings of the ACM International Conference on Multimedia},
  year={2023}
}

@inproceedings{erfba,
  title={Enhanced Receptive Field with Boundary-Aware Temporal Forgery Detection},
  author={Wang, Zhixi and Cai, Zhixi and Dhall, Abhinav},
  booktitle={Proceedings of the IEEE/CVF Conference on Computer Vision and Pattern Recognition Workshops},
  year={2023}
}

@article{ganfingerprint,
  title={{GAN} Fingerprints: Detecting and Attributing Fake Images to their Source},
  author={Yu, Ning and Davis, Larry and Fritz, Mario},
  journal={IEEE Transactions on Pattern Analysis and Machine Intelligence},
  volume={43},
  number={10},
  pages={3512--3524},
  year={2021}
}

@inproceedings{audiofingerprint,
  title={Half-Truth: A Partially Fake Audio Detection Dataset},
  author={Yi, Jiangyan and Fu, Ruibo and Tao, Jianhua and Nie, Shuai and Ma, Haoxin and Wang, Chenglong and Wang, Tao and Shao, Cunhang},
  booktitle={Interspeech},
  year={2022}
}

@inproceedings{resnet,
  title={Deep Residual Learning for Image Recognition},
  author={He, Kaiming and Zhang, Xiangyu and Ren, Shaoqing and Sun, Jian},
  booktitle={Proceedings of the IEEE Conference on Computer Vision and Pattern Recognition (CVPR)},
  pages={770--778},
  year={2016}
}

@inproceedings{focalloss,
  title={Focal Loss for Dense Object Detection},
  author={Lin, Tsung-Yi and Goyal, Priya and Girshick, Ross and He, Kaiming and Doll{\'a}r, Piotr},
  booktitle={Proceedings of the IEEE International Conference on Computer Vision (ICCV)},
  pages={2980--2988},
  year={2017}
}

@article{adamw,
  title={Decoupled Weight Decay Regularization},
  author={Loshchilov, Ilya and Hutter, Frank},
  journal={International Conference on Learning Representations (ICLR)},
  year={2019}
}

@inproceedings{facealignment,
  title={How Far are We from Solving the 2D \& 3D Face Alignment Problem? (and a Dataset of 230,000 3D Facial Landmarks)},
  author={Bulat, Adrian and Tzimiropoulos, Georgios},
  booktitle={Proceedings of the IEEE International Conference on Computer Vision (ICCV)},
  pages={1021--1030},
  year={2017}
}

@inproceedings{ffpp,
  title={FaceForensics++: Learning to Detect Manipulated Facial Images},
  author={Rossler, Andreas and Cozzolino, Davide and Verdoliva, Luisa and Riess, Christian and Thies, Justus and Nie{\ss}ner, Matthias},
  booktitle={Proceedings of the IEEE/CVF International Conference on Computer Vision},
  pages={1--11},
  year={2019}
}

@inproceedings{sst,
  title={Self-Supervised Transformer for Deepfake Detection},
  author={Zhao, Han and Zhou, Wei and Chen, Dongdong and Zhang, Weiming and Yu, Nenghai},
  booktitle={arXiv preprint arXiv:2203.01265},
  year={2022}
}

@inproceedings{avforensics,
  title={AVForensics: Audio-Driven Deepfake Video Detection with Masking Strategy},
  author={Zhu, Yinan and Gao, Jian and Zhou, Xi},
  booktitle={Proceedings of the 2023 ACM International Conference on Multimedia Retrieval},
  pages={162--171},
  year={2023}
}

@article{videofingerprint,
  title={Video Forensics: Identifying Video Manipulations},
  author={Verdoliva, Luisa},
  journal={IEEE Signal Processing Magazine},
  volume={39},
  number={1},
  pages={38--49},
  year={2022}
}

@inproceedings{DMASTA2022,
  title={Model Attribution of Face-Swap Deepfake Videos},
  author={Jia, Shan and Li, Xin and Lyu, Siwei},
  booktitle={2022 IEEE International Conference on Image Processing (ICIP)},
  pages={2356--2360},
  year={2022}
}

@inproceedings{celepdf,
  title={Celeb-DF: A Large-Scale Challenging Dataset for DeepFake Forensics},
  author={Li, Yuezun and Yang, Xin and Sun, Pu and Qi, Honggang and Lyu, Siwei},
  booktitle={Proceedings of the IEEE/CVF Conference on Computer Vision and Pattern Recognition},
  pages={3207--3216},
  year={2020}
}

@article{dolhansky2020,
  title={The Deepfake Detection Challenge (DFDC) Dataset},
  author={Dolhansky, Brian and Bitton, Joanna and Pflaum, Ben and Lu, Jikuo and Howes, Russ and Wang, Menglin and Ferrer, Cristian Canton},
  journal={arXiv preprint arXiv:2006.07397},
  year={2020}
}

@article{timit,
  title={Deepfakes and Face Swap: Detection and Generation},
  author={Korshunov, Pavel and Marcel, S{\'e}bastien},
  journal={arXiv preprint arXiv:1811.00656},
  year={2018}
}

@article{domainadapt,
  title={A Theory of Learning from Different Domains},
  author={Ben-David, Shai and Blitzer, John and Crammer, Koby and Kulesza, Alex and Pereira, Fernando and Vaughan, Jennifer Wortman},
  journal={Machine Learning},
  volume={79},
  number={1},
  pages={151--175},
  year={2010}
}

\end{document}